\documentclass[final,3p,times,twocolumn]{elsarticle}

\usepackage{soul}
\setstcolor{red}
\usepackage{xcolor}
\usepackage{ulem}

\usepackage{amssymb}
\usepackage{amsmath}
\usepackage{orcidlink}
\usepackage{graphicx}
\usepackage{caption}
\usepackage{subcaption}
\usepackage{todonotes}
\usepackage{float}
\usepackage{url}
\usepackage{xurl}
\usepackage{booktabs}
\usepackage{multirow}
\usepackage{tabularx}

\newcolumntype{Y}{>{\centering\arraybackslash}X}

\journal{Environmental Modeling and Software}

\begin{document}
\let\WriteBookmarks\relax
\def\floatpagepagefraction{1}
\def\textpagefraction{.001}

\begin{frontmatter}



\title{Time Distributed Deep Learning Models for Purely Exogenous Forecasting: Application to Water Table Depth Predictions using Weather Image Time Series}

%\author{} %% Author name

\author[aff1,aff2]{Matteo Salis\corref{cor1} \orcidlink{0009-0009-2810-9992}} %
\ead{matteo.salis@unito.it}
\ead[url]{https://sites.google.com/view/matteo-salis/home}

\author[aff2]{Abdourrahmane M. Atto \orcidlink{0000-0003-1753-4917}}
\ead{abdourrahmane.atto@univ-smb.fr}

\author[aff3]{Stefano Ferraris \orcidlink{0000-0001-8544-6199}}
\ead{stefano.ferraris@polito.it}

\author[aff1]{Rosa Meo \orcidlink{0000-0002-0434-4850}}
\ead{rosa.meo@unito.it}

%\cormark[1]



%\credit{Conceptualization of this study, Methodology, Software}

\affiliation[aff1]{organization={Computer Science Department - University of Turin},
            addressline={Corso Svizzera 185}, 
            city={Torino},
            postcode={10149},
            country={Italy}}

\affiliation[aff2]{organization={LISTIC Laboratory - Université Savoie Mont Blanc},
            addressline={5 chemin de bellevue}, 
            city={Annecy-le-vieux},
            postcode={74 940}, 
            country={France}}

\affiliation[aff3]{organization={Interuniversity Department of Regional and Urban Studies and Planning, Politecnico di Torino and University of                 Turin},
            addressline={Viale Pier Andrea Mattioli 39}, 
            city={Turin},
            postcode={10125}, 
            country={Italy}}

\cortext[cor1]{Corresponding author}

%% Abstract
\begin{abstract}
Deep Learning (DL) models have revealed to be very effective in hydrology, especially in handling spatially distributed data (e.g. raster data). We have proposed two different DL models to predict the water table depth in the Grana-Maira catchment (Piedmont, IT) using only exogenous weather image time series. Both the models are made of a first Time Distributed Convolutional Neural Network (TDC) which encodes the images into hidden vectors. The first model, TDC-LSTM uses then a Sequential Module based on an LSTM layer to learn temporal relations and output the predictions. The second model, TDC-UnPWaveNet uses instead a new version of the WaveNet architecture, adapted for handling output of different length and completely shifted in the future to the input. Both models have shown remarkable results focusing on different learnable information: TDC-LSTM has focused more on bias while the TDC-UnPWaveNet more on the temporal dynamics maximizing correlation $\rho$, achieving mean BIAS (and standard deviation) -0.18(0.05), -0.25(0.19) and $\rho$ 0.93(0.03), 0.96(0.01) respectively over all the sensors. 
\end{abstract}

\begin{keyword}
% You are required to provide 1 to 7 keywords
Deep Learning \sep Image Time Series \sep CNN \sep LSTM \sep WaveNet \sep Groundwater Resources \sep Water Table Depth
\end{keyword}

% \begin{highlights}
% \item Deep learning is effective for water table depth predictions
% \item Accurate water table depth predictions are feasible just using exogenous weather data
% \item Deep learning Time-Distributed models can be used to handle image time series
% \item UnPWaveNet model is proposed as a new adaptation of the WaveNet architecture
% \end{highlights}

\end{frontmatter}

% Main text
\section{Introduction}\label{sec:intro}
%State the objectives of the work and provide an adequate background, avoiding a detailed literature survey or a summary of the results
Water management is a key element in the framework of sustainable development, even more so in the context of climate change~\cite{IPCCClimate2023}. 
Freshwater is essential for sanitation and hygiene standards, but also for food availability and economic stability~\cite{EEAWater2023}. 
Groundwater resources are the second most important component of the hydrological cycle, accounting for about 30\% of the world's freshwater resources, just behind glaciers and ice caps, which account for 68.7\%~\cite{ShiklomanovWorld1998}. 
Moreover, groundwater resources prove to be a more stable source of freshwater compared to rivers and lakes, whose water storage varies promptly with the current weather conditions~\cite{FamigliettiGlobal2014}. 
Indeed, in Europe, 65\% of drinking water and 25\% of irrigation comes from groundwater resources~\cite{EEAEurope2022}; these percentages are expected to increase given the expected effects of climate change~\cite{IPCCClimate2023,EEAWater2021,EEAWater2023,JasechkoRapid2024}.\\
Water management policies are a key tool to pursue and achieve sustainable development~\cite{UNUnited2023,UNGlobal2023} and to meet water needs both in terms of drinking water for humans and irrigation for crops.
Thus, accounting for water resources is paramount for planning water policies, and it is essential to develop models that can quantify groundwater resources. \\
The traditional process-based approaches consist of developing hydrological models mapping explicitly the cause-effect relationship among the variables under study.
Different solutions have been proposed ranging from conceptual to physically-based and distributed models~\cite{Sabzipour2023,Tripathy2024}.
However, calibrating process-based model parameters requires many geophysical data of the aquifer under study and a deep knowledge of the hydrogeophysical processes involved~\cite{KirchnerDouble2003,BloschlTwentythree2019}.
This turns into costly and time-consuming field measurements, that in some developing rural areas are even unfeasible~\cite{KauffeldtDisinformative2013,BolsterRecent2019,ChenComparative2020,YinComparison2021}. 
Furthermore, even if computing resources have grown substantially, they are still stretched to their limits by simulation of complex spatio-temporal hydrological phenomena~\cite{Clark2017}.
Indeed, execution times of simulations remain an issue hindering the use of accurate process-based models for nowcasting applications.
Nevertheless, process-based models remain a cornerstone for developing explainable models that obey physical laws, and in cases in which the possibility of explaining how the model produces a specific output (i.e. having a white box model) is more relevant than having accurate and fast predictions. 
Authors in~\cite{He2020mathappr} proposed a new mathematical method to improve the modeling of surface–groundwater interactions while still maintaining a formal description of the model functioning.
In more detail, they adopted spectral analysis to improve the accuracy of the Fourier fitting process to identify the significant surface signals. The approach was tested on synthetic data showing improved accuracy in approximating surface and boundary fluxes in the hydrological model.\\
In the last years, different, and fully empirical, approaches have emerged leveraging the growing availability of hydrological measurement data.
In more detail, machine learning techniques have been applied to groundwater modeling and have proven to be very efficient in achieving remarkable results without any prior domain knowledge, neither in terms of physics laws nor in terms of aquifer geophysical details~\cite{Zounemat2021ensml,TaoGroundwater2022,May-LagunesForecasting2023}. Deep Learning (DL) models, based on neural networks, have reached even better performances thanks to their ability to learn very complex and nonlinear relations between input and output~\cite{CoulibalyArtificial2001,MohantyComparative2013,YinComparison2021,ClarkForecasting2022,Sabzipour2023,Tripathy2024}.
Indeed, neural networks have been demonstrated to be universal approximators~\cite{Hornik1989}, and thus they can approximate any continuous function with a sufficient number of model parameters and an adequate amount of data. 
A critical aspect of deep learning is the computational cost of training neural networks. However, highly optimized libraries and Graphical Processing Units (GPU) that speed up the execution of parallelized algorithms help in reducing training times. Furthermore, it is worth considering that once a model is trained, predictions could be obtained rapidly, promoting the adoption of DL for nowcasting~\cite{Shi2017nowcasting}.\\
Looking more at the implementation viewpoint, DL models for groundwater application have been developed by using mainly exogenous weather data as input, and in some cases, some autoregressive terms (past target data) have been added as input to improve performances.
% The phenomena related to groundwater are extremely complex and strongly related to the physical properties of the context~\cite{KirchnerDouble2003,BloschlTwentythree2019}, not by chance physical models need a lot of information about the region of interest and are very scale dependent~\cite{BolsterRecent2019,KauffeldtDisinformative2013}.
% In recent years, Deep Learning (DL) techniques have been applied to groundwater modeling and have proven to be very efficient in achieving remarkable results without requiring all the physical knowledge of the region under study~\cite{CoulibalyArtificial2001,MohantyComparative2013,ZhangDeveloping2018,ChenComparative2020,YinComparison2021,ClarkForecasting2022,TaoGroundwater2022,May-LagunesForecasting2023}.
% In particular, in addition to autoregressive terms providing information on the past states of the variable to be predicted (target variable), most DL studies use exogenous weather variables (e.g. precipitation, temperature, etc.) as the main input, motivated by domain knowledge and data availability. 
Other types of data, such as anthropogenic pressure on water resources (e.g. water abstraction, irrigation), are only used by a minority of studies due to their scarce availability in many catchments and at large scales~\cite{DollGlobalscale2014,LeeUsing2019,ClarkUnravelling2022}.\\
To quantify groundwater resources, groundwater level (GWL) is frequently adopted as the target variable~\cite{TaoGroundwater2022}. GWL represents the distance between the reference datum and the water table, which in turn is defined as the higher surface of the groundwater body. Another possibility of quantification is to measure the water table depth, i.e. the distance between the ground surface and the water table. 
In this way, a decrease in the depth of the water table means an increase in groundwater resources; conversely, an increase in the depth of the water table means a decrease in the amount of water stored in the phreatic aquifer. 
The more widely used data type for groundwater depth prediction is tabular data (i.e. one-dimensional time series) for both input and output, as in~\cite{WunschGroundwater2021,ClarkForecasting2022,May-LagunesForecasting2023}.
However, the authors of~\cite{WunschKarst2022,AndersonEvaluation2022} demonstrated the usefulness of using spatially distributed input data (e.g. raster data) as input data for hydrological data-driven modeling, allowing the DL model to find the most useful relationships among all input variables distributed over the region of interest (ROI). 
This could be done by taking as input a time series of weather raster images (i.e. a multidimensional time series), where for each weather variable an image is retrieved at each time step $t$.
The values of a pixel in an image are the vector of values associated with the corresponding weather variables for the area covered by this pixel.\\
Spatially distributed data (i.e. images) could be efficiently and effectively handled by 2D Convolutional Neural Networks (CNNs), which have been extensively and proficiently adopted in DL models to deal with images and learn spatial relationships~\cite{KrizhevskyImageNet2012,SzegedyGoing2014,ZeilerVisualizing2014,HeDeep2015,FerrarisMachine2023}. CNNs have been explored also for learning temporal relationships, i.e. sequential data; for example, 1D CNNs have been widely adopted~\cite{LeCunConvolutional1995} also in water resources studies~\cite{WunschGroundwater2021,AliFlood2022,TaoGroundwater2022}. 
However, Recurrent Neural Networks (RNNs) and Long-Short-Term Memory (LSTM) have proven to be very hard to beat competitors for sequential data, thanks to their ability to learn both short and long term dependencies and their resilience to noise~\cite{HochreiterLong1997,GersLearning2002}. 
In recent years, many researchers have attempted to build new and more sophisticated CNN-based architectures to compete with LSTM on sequential data~\cite{LeaTemporal2016,GehringConvolutional2017,BaiEmpirical2018,BorovykhConditional2018,LiConvolutional2018,BaiTrellis2019,IsmailFawazInceptionTime2020,BrockInvestigating2023}. 
These strengths are justified by the fact that CNNs are less computationally intensive and highly parallelizable, meaning that their training is 
less energy-consuming. 
Furthermore, with the use of dilated convolution (also called "hole convolution")~\cite{YuMultiScale2016}, it has been possible to implement CNN-based models with the ability to capture very long-term relationships with a smaller number of parameters.
Different architectures have been proposed adopting the dilated convolution, and they have shown very competing performances in comparison to RNN~\cite{LeaTemporal2016,BaiEmpirical2018}. One example could be the WaveNet model, developed by Google~\cite{OordWaveNet2016} for audio generation in a many-to-many framework\footnote{In machine learning literature there are different frameworks for sequential data modeling: many-to-one (or seq2one) for models which take as input a sequence and output a scalar, many-to-many (or seq2seq) for models which take as input a sequence and output a sequence.}. WaveNet uses dilated convolution to learn long dependencies and employs causal padding to constrain each element of the output sequence to depend only on past input observations (see Section~\ref{sec:methods} for more details). This architecture has been widely used for tasks other than audio generation, with remarkable results~\cite{BorovykhConditional2018,DoyleAccurate2020,BrockInvestigating2023}.\\
To deal with image time series many studies combined 2D CNN and a sequential model (e.g. LSTM or 1D CNN) to get the most from them. 
More specifically, 2D CNNs have been used in a time-distributed (TD) manner, i.e. the same 2D CNN is applied to the image available at each time step, and spatial relationships are extracted from it. The output of the TD 2D CNN is then fed into a sequential model that focuses on the temporal relationships of the data - sometimes these models are referred to in the literature as hybrid models (e.g. CNN-LSTM)~\cite{ChenHybrid2019,ZangShortterm2020,CastroSTConvS2S2021,DingHybrid2021,YanMultihour2021,AggaCNNLSTM2022,AndersonEvaluation2022,WunschKarst2022}.
As mentioned above, from a statistical viewpoint, an image time series of many weather variables can be seen as a multidimensional (in the dimensions of time, longitude and latitude) and multivariate (more than one input variable) time series.
Instead, from a Computer Science point of view, this data flow could be seen as a video in which each frame (i.e. time step) contains many channels (i.e. variables or features). Indeed, many studies adopt these hybrid models for different video tasks~\cite{UllahAction2018,AbdullahFacial2020,YangCNNLSTM2020}. 
To be readable by the two communities, in the following, we will use the term channels interchangeably with variables, and image time series with video.\\
Regarding the application, the present research focuses on the Grana-Maira catchment in Piedmont (Italy).
Our objective, a many-to-one task,  is to predict the weekly water level depth measured by three sensors in the catchment area (i.e. our ROI). The input consists of exogenous weather information from the last two years and takes the form of image time series over the ROI.
To this aim and inspired by literature, we developed and compared two different composite (i.e. made of sub-modules) DL models. 
The first sub-module, named Time Distributed CNN (TDC) is the same for both, it is responsible for handling the images available at each time step of the time series while learning spatial relations.
What distinguishes the two models is the second module (hereafter Sequential Module), which is responsible for learning temporal relations. 
The first model, named TDC-LSTM, has a Sequential Module based on an LSTM layer.
Differently, the second model, named TDC-UnPWaveNet, uses a new version of the WaveNet, which is proposed here to be usable for the many-to-one case and, in general, for tasks in which the output sequence has a lower length and it is completely shifted in the future to the input sequence.
This adaptation required a restructuring of the original WaveNet model and the development of a new Channel Distributed (CD) layer to handle objects of different time lengths between the hidden layers of the architecture.

\subsection{Related Works}\label{sec:intro_rw}

DL techniques have already been applied proficiently to groundwater resource forecasting.
In \cite{WunschGroundwater2021} authors made a comparison of different DL architectures, namely NARX \cite{LinLearning1996}, 1D CNN, and LSTM to predict GWLs on 17 sensors in the Upper Rhine Graben (URG) region. NARX is a neural network architecture specifically designed to model autoregressive terms (i.e. the past values of the target) and exogenous data in a nonlinear fashion.
They trained local models for each sensor using Bayesian optimization and fed as input weather variables measured by nearby sensors (in other words inputs had a tabular structure); furthermore, for some of the sensors, there was also an autoregressive term.
The authors found that CNN was faster in training and inference than other methods.
LSTM was revealed to be the worst performing, while NARX performed the best, but this conclusion could be due to the intrinsic use of an autoregressive component by the NARX, not explicitly provided to CNN and LSTM.
Nevertheless, in many other works, LSTM performed better than other methods like ARIMA, ANN~\cite{SunDatadriven2022}, and Random Forest~\cite{YinComparison2021}. In~\cite{JeongComparative2019a}, the authors made a comparison across many models and found that LSTM and NARX were the best models, with no clear winner between the two. However, LSTM seems to be more established and widespread in the DL community and also in hydrological applications, which more often adopt LSTM as a reference model for data-driven modelling~\cite{HuDeep2018,ShenTransdisciplinary2018,XiangRainfall2020,NouraniDeep2022}.\\
Most of the studies on groundwater modeling deal with tabular data to model their target. This means retrieving input and output data directly from the measurement sensor network over the region of interest (i.e. geospatial data in more statistical terms).
In this way, each observation in the dataset is indexed by the time of acquisition, and longitude and latitude of the sensor by which it is measured.
Spatially distributed data instead are represented in a grid format (i.e. raster) in which each portion of the area under study is represented by a square of the grid (or a pixel if one looks at the raster as an image).
This type of data represents the variables of interest all over the region under study, and not only on the coordinates in which the sensors are located.
Spatially distributed data could be created either by spatial interpolation of tabular data or by using other sensors like satellites (e.g. GRACE data~\cite{CSR2021}).
In both cases, in the DL framework, using spatially distributed data as input could facilitate the model in understanding spatial relationships and extracting the most relevant information without making a priori assumptions on the form of these spatial relations. \\
Some works in hydrology retrieved spatially distributed data, however, many of these have reduced the spatial dimension before feeding the data into the model, for example by averaging the value of the variable over the watershed of interest~\cite{AmarantoSemiseasonal2018,May-LagunesForecasting2023}. 
Instead, the authors in~\cite{AndersonEvaluation2022,WunschKarst2022} used spatially distributed weather data (i.e. raster images) as direct input to model streamflow in Canada and spring discharge in karst catchments, respectively.
In general, the use of spatially distributed data could be helpful for two reasons.
The first is that, thanks to new open raster datasets such as ERA5-land~\cite{Munoz-SabaterERA5Land2021}, it is possible to retrieve water and energy-related variables even in regions that do not have an extensive sensor network.
The second is the ability to build complex models that autonomously learn the most relevant spatial relationships between the weather input variables and the target.
In more detail, in \cite{AndersonEvaluation2022} a neural network made by 2 modules is proposed to jointly predict streamflow for 226 stream gauge stations.
The first module is a TD CNN made of 5 convolutional layers and 2 max pooling layers. The input images at each time step are squeezed in the spatial dimension to obtain a vector of dimension 32, then the input video is transformed into a multivariate (hidden) time series with 32 features. The second module is a one-layer LSTM with 80 units, and the last is a fully connected layer that outputs the predictions for the 226 stream gauges.
In \cite{WunschGroundwater2021}, the authors made a comparison between the tabular and spatially distributed approaches in three different areas. For the spatially distributed data, they still developed a 2-module model, where the first module is a TD CNN, and the second module is a 1D CNN layer.
They argued that by using spatially distributed data, one can overcome the difficulties in areas with a low density of sensors. Furthermore, for studies focused on more catchments, and with the aid of sensitivity analysis, it is possible to extract a naive localization of the catchment by looking at the most sensitive pixels in the original raster images - a practice belonging to the perturbation methods in the eXplainable AI (XAI) research field~\cite{AndersonEvaluation2022,PetsiukRISE2018}.\\
To the best of our knowledge, no other studies have attempted to feed spatially distributed data directly into a DL model to predict the water table depth, especially in our ROI, and to make a direct comparison between hybrid recurrent and convolutional methods. 
Thus, our contributions are the following: 
\begin{enumerate}
    \item Development of DL models for water table depth predictions in the Grana-Maira catchment in Piedmont (IT) 
    \item Development and application of DL hydrological models that directly use spatially distributed data 
    \item Adaptation of WaveNet to the many-to-one case through UnPWaveNet, a new competitor to recurrent architectures for sequential data.
\end{enumerate}

\subsection{Case Study Description}\label{sec:intro_cs_data}
\begin{figure*}[h]
	\centering
	\includegraphics[width=.9\textwidth]{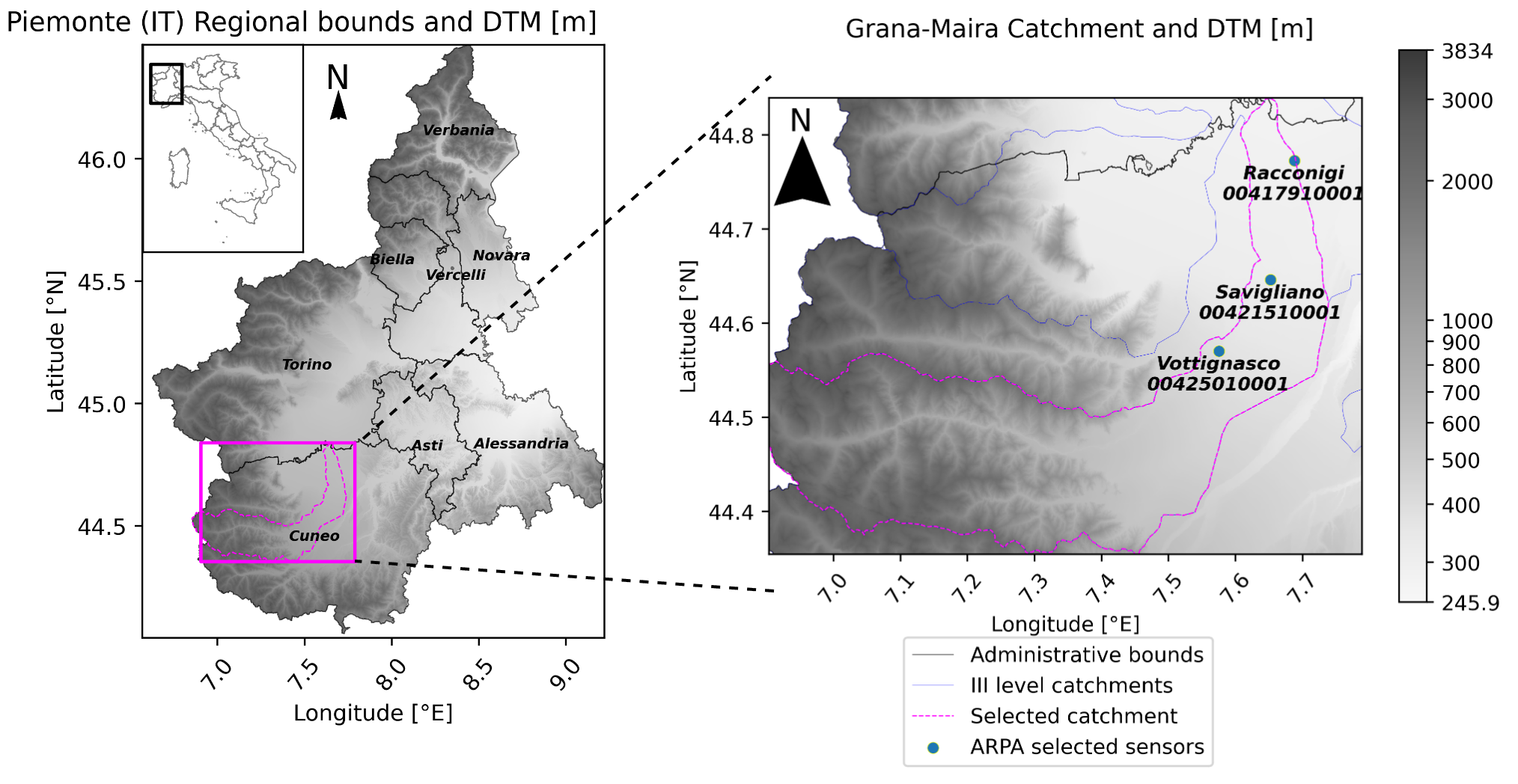}
	\caption{Piedmont Region with administrative Province (bold names). Zoom on the Grana-Maira selected catchment (in magenta) with the three water table sensors. The digital elevation model (DTM) is represented in meters [m] through a grey logarithmic scale.}
	\label{fig:roi_dtm}
\end{figure*}

Groundwater resources in Italy are even more exploited than the European average.  Indeed, 85\% of drinking water in Italy comes from groundwater sources~\cite{ISTATStatistiche2023}. 
In Piedmont, an administrative region in the north-west of Italy, almost half of the total water abstracted is used by the agricultural sector, which makes extensive use of irrigation to meet the water needs of crops~\cite{RegionePiemontePTA2021,AdBDPoPiano2021}. 
It is estimated that 83\% of irrigable land is effectively irrigated, a fact that indicates the limited use of seasonal rainfall as a direct source of water instead of human abstractions~\cite{CREAAgricoltura2023}.\\
Piedmont is a very heterogeneous region from the geographical point of view, with Alps near the western, northern, and southern borders, hills extending from the center to south-est, and plains that cover an area from the Cuneo Province in the south-west to the upper-central part in Vercelli and Novara Provinces.
In this context is very difficult to analyze groundwater resources, also because of the intensive agricultural activities. 
Indeed, the authors in~\cite{BrussoloAquifer2022} analyzed aquifer recharge in the Piedmont Alpine zone in the north-west and they highlighted the difficulty of finding a general trend between different areas, in other words, the results of their analysis are very context-dependent.
For this reason, we decided to focus on a specific catchment called Grana-Maira, located in the Cuneo Province in the southwest of Piedmont (Figure~\ref{fig:roi_dtm}). 
The Grana-Maira basin takes its name from the two rivers (Grana and Maira) that originate in the Alps in the western part of the basin.
The altitude of the catchment area decreases from west to northeast, i.e. from mountain to plain.

\subsubsection{Data}
\begin{figure*}
\centering
    \begin{subfigure}[c]{0.85\textwidth}
        \centering
        \includegraphics[width=\textwidth]{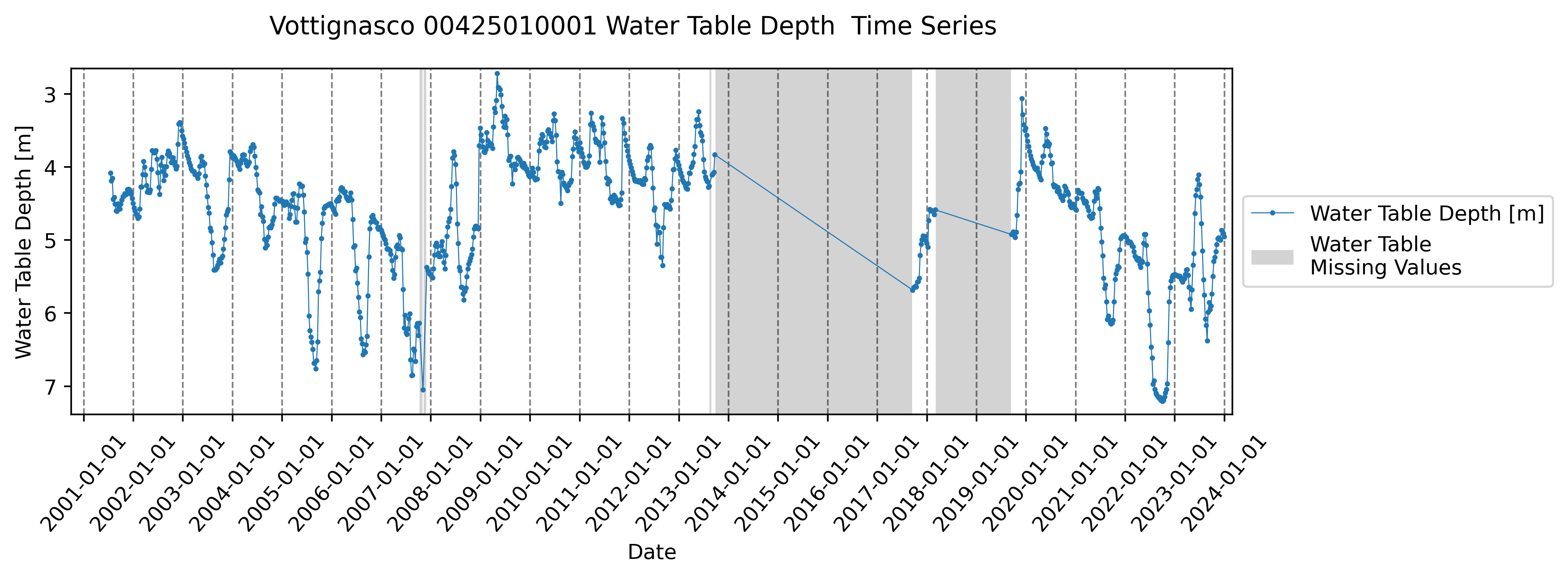}
        \caption{}
        \label{fig:vottignasco_ts}
    \end{subfigure}
    %\hfill
    \begin{subfigure}[c]{0.85\textwidth}
        \centering
        \includegraphics[width=\textwidth]{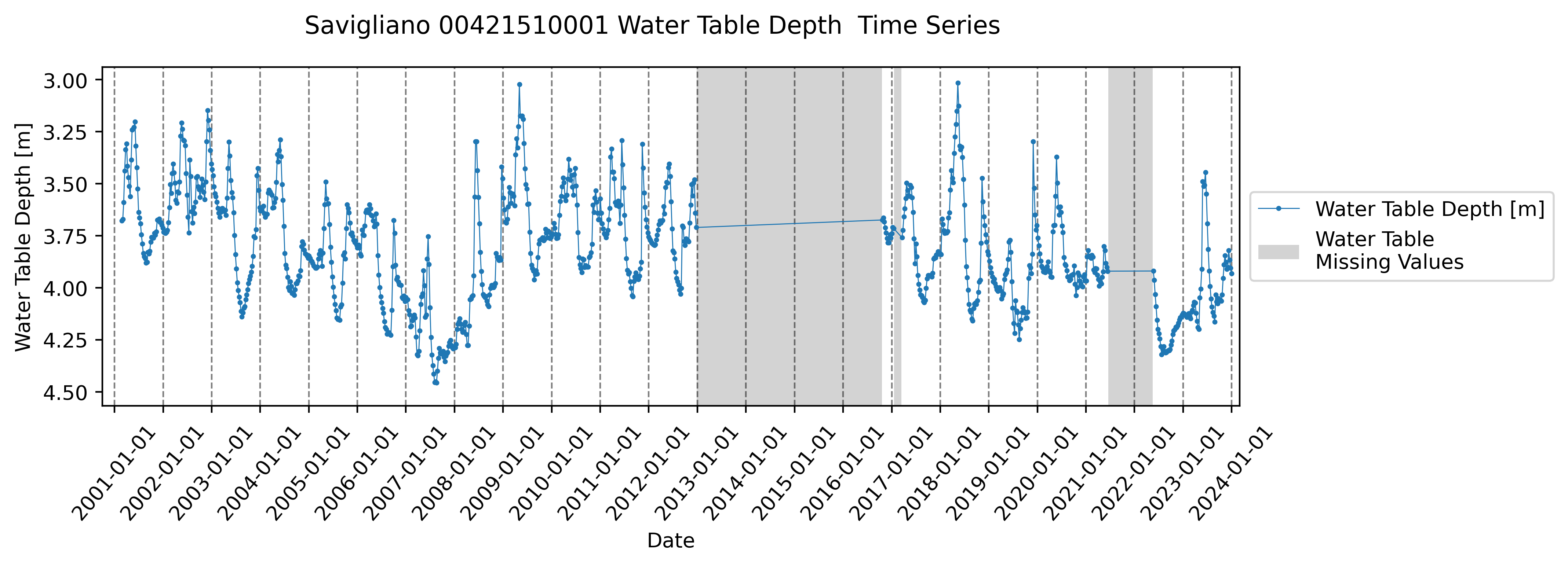}
        \caption{}
        \label{fig:savigliano_ts}
    \end{subfigure}
    \begin{subfigure}[c]{0.85\textwidth}
        \centering
        \includegraphics[width=\textwidth]{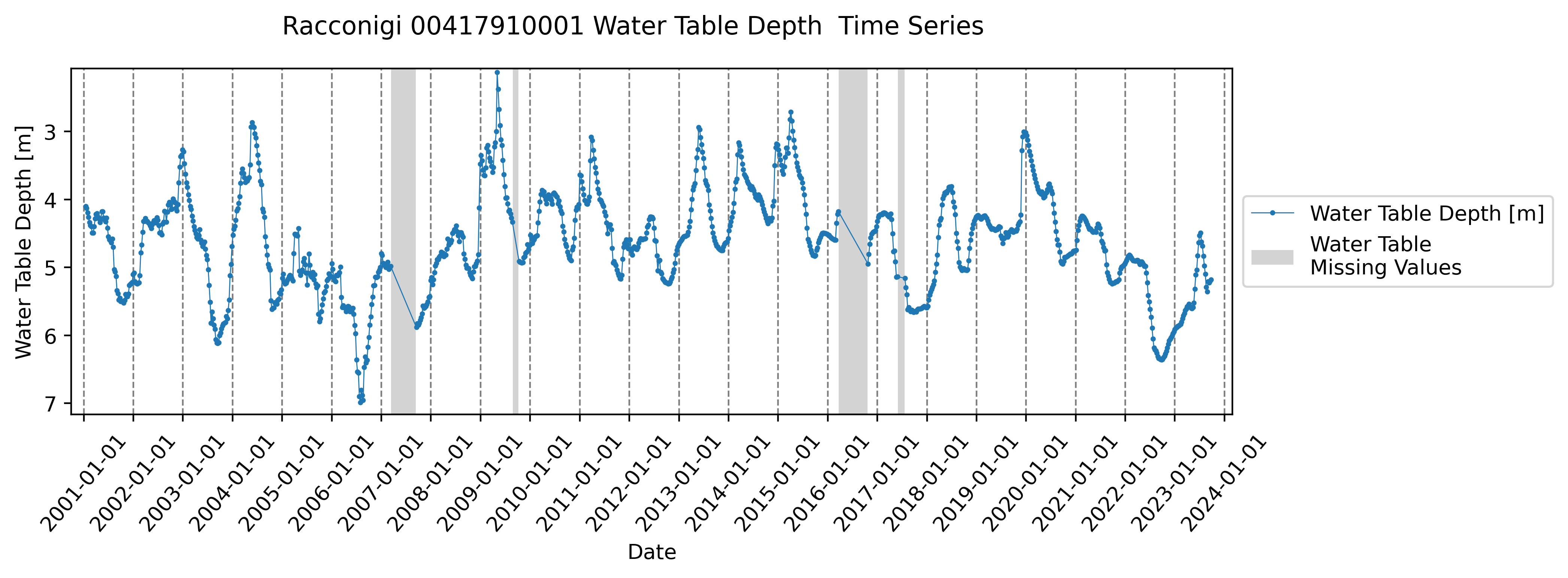}
        \caption{}
        \label{fig:racconigi_ts}
    \end{subfigure}
\caption{Water Table Depth weekly time series measured by sensors \textbf{a)} Vottignasco 00425010001, \textbf{b)} Savigliano 00421510001, and \textbf{c)} Racconigi 00417910001. All the plots have the y-axis reversed (lower value on top) because in this way it should be easier to interpret the water table depth: if it increases it means a decrease in water resources stored in the ground.}
\label{fig:wt_series}
\end{figure*}

We retrieved three time series of the water table depth from sensors in the municipalities of Vottignasco, Savigliano, and Racconigi. These sensors are part of the measurement network of the Regional Environmental Agency (ARPA\footnote{In Italian Agenzia Regionale per la Protezione Ambientale.}) and are freely available by request. Table~\ref{tab:wt_statistics} reports different information and summary statistics for the three series, and Figure~\ref{fig:wt_series} shows the weekly average series for the three sensors. This figure highlights a problem of missing data and irregularities between the series. Indeed, some huge missing periods are present and, furthermore, these gaps are not the same for the three series. 
Given the large extension of some of these missing periods, it was considered more sensible not to perform imputation, but to let the deep learning models learn from the available data.\\
For the present work, weather raster data of total precipitation, maximum, and minimum temperature at 0.125° spatial resolution were retrieved from \cite{ARPADataset} and solely used as the input of the proposed models. 
We decided to not include an autoregressive component (i.e. past values of water table depth as an additional input) because groundwater data are released on a semester basis, then using an autoregressive component would make the proposed models unusable at present time because of the lack of the recent water table data.
Instead, weather data are updated daily without any missing values. Thus, for each water table depth point to be predicted, it is possible to construct a video made of frames each one with three channels, namely total precipitation, and maximum and minimum temperature.

\begin{table*}[t]
\centering
\caption{Water Table Depth time series statistics. The sensor column shows the municipality and the identification codes of the sensors; $\mu$ represents the global mean; $\sigma$ represents the global standard deviation; Observation represents the total number of observations; First and Last data represent the dates of the first and last measurement respectively.}
\begin{tabularx}{\textwidth}{c|YYYYY}
\toprule
\textbf{Sensor} & \textbf{$\mu$ [m]} & \textbf{$\sigma$} [m] & \textbf{Observations} & \textbf{First date} & \textbf{Last date} \\ 
\hline\hline
\textbf{Vottignasco 00425010001} & 4.43        & 0.75 & 879 & 2001-07-15 & 2023-12-31                               \\ \hline
\textbf{Savigliano 00421510001}  & 3.77        & 0.27 & 938 & 2001-02-25 & 2023-12-31                    \\ \hline
\textbf{Racconigi 00417910001}   & 4.54        & 0.76 & 1116 & 2001-01-14 & 2023-09-24                             \\
\bottomrule
\end{tabularx}
\label{tab:wt_statistics}
\end{table*}

\section{Methods}\label{sec:methods}

\begin{figure*}[t]
	\centering
	\includegraphics[width=.95\textwidth]{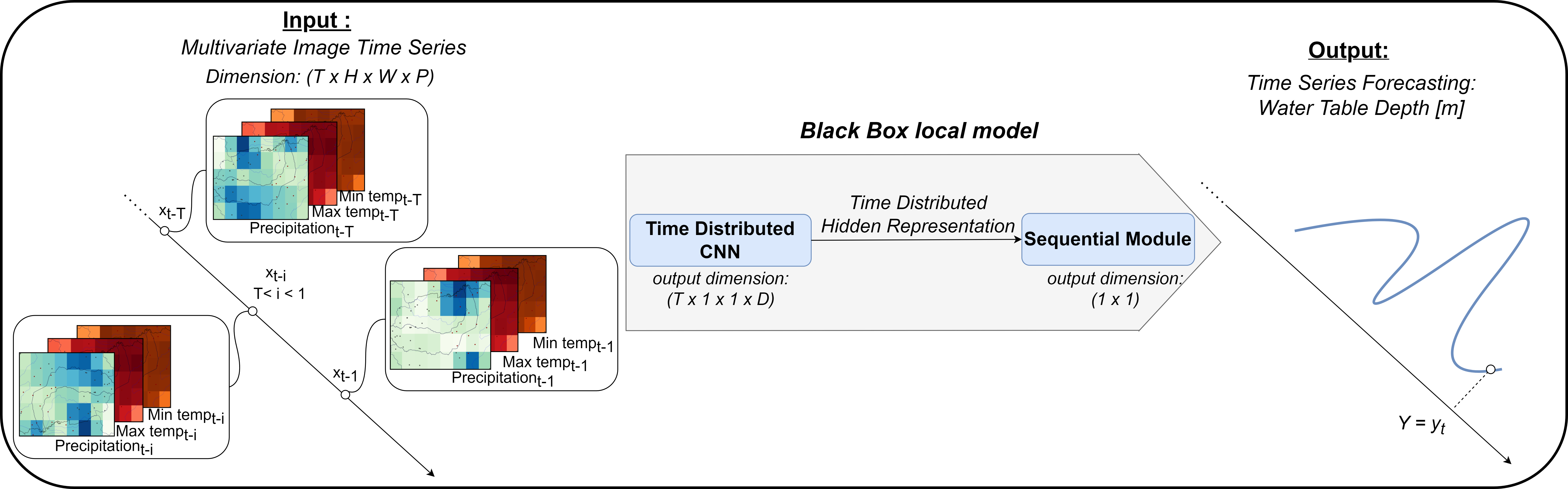}
	\caption{Proposed modeling pipeline with the general structure of the black box local models.}
	\label{fig:pipeline}
\end{figure*}

This work has aimed to train local models for each sensor independently of the others. This is because it is consistent with the literature (see for example \cite{WunschGroundwater2021,AndersonEvaluation2022,May-LagunesForecasting2023}) but also because of the irregularities between the series explained in Section~\ref{sec:intro_cs_data}. 
Indeed, if one wanted to train a global model using the data from all the sensors, it would require getting only the data from non-missing dates available jointly to all three sensors, discarding in this way a lot of information. Furthermore, as shown in Section~\ref{sec:intro_cs_data}, the three series show different dynamic behavior and a general global model could be too restrictive and of scarce utility for the domain application.\\
Our proposed modeling pipeline is described in Figure~\ref{fig:pipeline}, and it consists in black box local models that take as input a multivariate and multidimensional time series, i.e. a video $X_t^{H,W,P,T} = \{
x(h;w;p;\tau) \in \mathbb{R}\, |\,h\in[1;H], w\in[1;W], p\in[1;P], \tau\in[t-T+1;t]\}$, of spatial extent $H \times W$ (i.e. height and width of each frame), time length $T$, and $P$ weather features. Each local model forecasts the water table depth at a weekly time step $t$ in a sliding window fashion.  
Formally, a local model has to learn the relation $f$ which links $y_t = f(X_{t-1}^{H,W,P,T}) + \epsilon_t$, where $\epsilon$ is the irreducible error term.
Given that groundwater phenomena $y$ could have a very long memory, we have used a very long input weather image time series length, setting T to 104 (i.e. 2 years in weekly terms) and letting the DL models use the most relevant past information. \\
The general structure of the proposed black box models consists of two modules (Figure~\ref{fig:pipeline}). The first module is a Time Distributed CNN (TDC) whose structure is identical to the two models. 
The TDC is responsible for learning spatial relations and it outputs a Time Distributed Hidden Representation of the input image time series, i.e. it encodes each frame of the video into a vector of dimension $D$; in other words, it extracts a classical multivariate (of $D$ variables) time series from the input.
The second module, the Sequential Module, is what characterizes the two different models called TDC-LSTM and TDC-UnPWaveNet.
In the following, the building boxes of the two models are explained, especially the modification and novelties carried out in the development of the UnPWaveNet, as the Channel Distributed layer. 

\subsection{Distributed Layers}
\subsubsection{Time Distributed Layers}
\label{sec:methods_td}

Generally speaking, inside a standard Neural Network, a hidden layer is made of many neurons, each of which is responsible for computing an affine transformation of the input and applying a non-linear activation function $g$.
Formally $A_{k,l}=g(W_{k,l}^\intercal A_{l-1} + \beta_{k,l})$ where $A_{k,l}$ is the output of a general neuron $k$ in layer $l$, and $A_{l-1}$ is the output matrix of the layer $l-1$; $W$ and $\beta$ contain the parameters to be learned.
It is possible to substitute the "simple" neuron with more complex operations, still made of neurons, but encapsulated in a so-called cell, while the formal neurons inside the cell are referred to as "units".
An exemplification concerning sequential data could be RNNs, and in particular, the LSTM layer, which is made of as many cells as the number of elements in the input sequence (or time step in the case of temporal data). Every cell is responsible for extracting long and short-term temporal dependencies and passing this information to the subsequent cell (through recurrent connection) \cite{HochreiterLong1997,GoodfellowDeep2016}. However, in the LSTM layer, and in RNN in general, the weights used by neurons in a cell are the same for every cell in the layer.
In other words, each element of the input sequence is processed by a cell with the same parameters, what differs between the cells of an LSTM layer are the inputs of every cell. \\
The concept of applying the same set of operations to every element of the input sequence is not applied only in RNNs, but it is a general way of proceeding also in other architectures.
A layer that works in this way is usually referred to as a Time Distributed (TD) layer.
Not by chance, in Keras\footnote{Open Source Python library for developing neural network models \url{https://keras.io/}.} there is a specific layer-class named \texttt{TimeDistributed} which applies the same set of operations to each element of the input sequence.
Figure~\ref{fig:td_layer} represents a general TD layer made of a simple fully connected cell applied on a multivariate time series $Z$ of length $T$ with $C$ variables, i.e. $Z = \{Z_\tau \in \mathbb{R}^C |\ \tau \in [t-T, t-1]$\}.
In Figure~\ref{fig:td_layer} the TD layer acts time-step by time-step transforming each vector $Z_\tau \in \mathbb{R}^C$ into a vector $Z'_\tau \in \mathbb{R}^{C^*}$, in which $C^*$ is defined by the number of neurons of the fully connected cell.
If $C^* > C$ the TD layer will dilate the channel dimension of each time step, reversely if $C^* < C$ (as in Figure~\ref{fig:td_layer}) the TD layer will squeeze the channel dimension.
It is relevant to point out that with a TD layer, the ordering and the length of the input sequence $Z$ are untouched and maintained also in the output sequence $Z'$. 
It is possible to develop TD layers with other types of cells than fully connected, and, as we already discussed, one of the most used TD networks for multidimensional sequences (e.g. video) is the TD CNN \cite{UllahAction2018,AbdullahFacial2020,HuTemporally2020a}.
In the case of TD CNN, the same CNN is applied on each frame of the video extracting spatial features. In Section~\ref{sec:methods_models_TDCmodule} we describe the implementation of our TDC module based on TD CNN.

\subsubsection{Channel Distributed Layers}
\label{sec:methods_cd}

\begin{figure*}[t!]
\centering

    \begin{subfigure}[c]{0.9\textwidth}
        \centering
        \includegraphics[width=\textwidth]{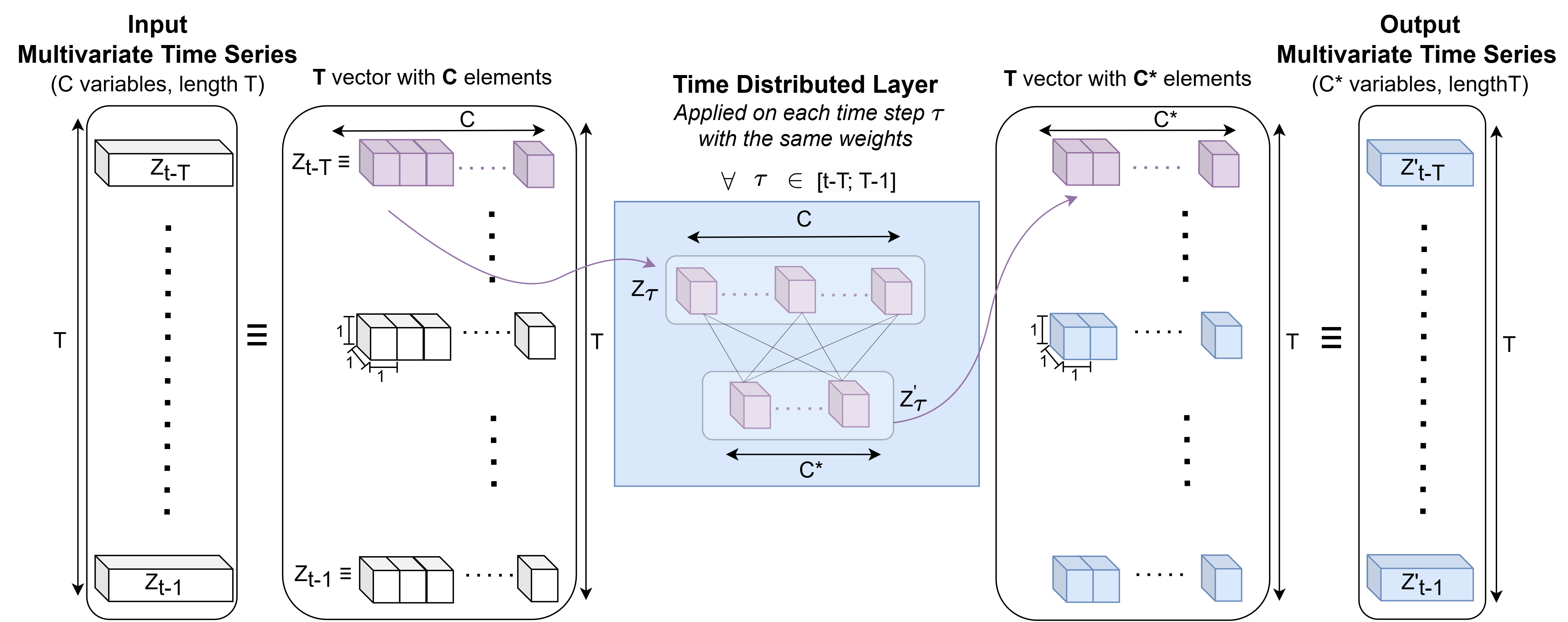}
        \caption{}
        \label{fig:td_layer}
    \end{subfigure}
    %\hfill
    \begin{subfigure}[c]{0.9\textwidth}
        \centering
        \includegraphics[width=\textwidth]{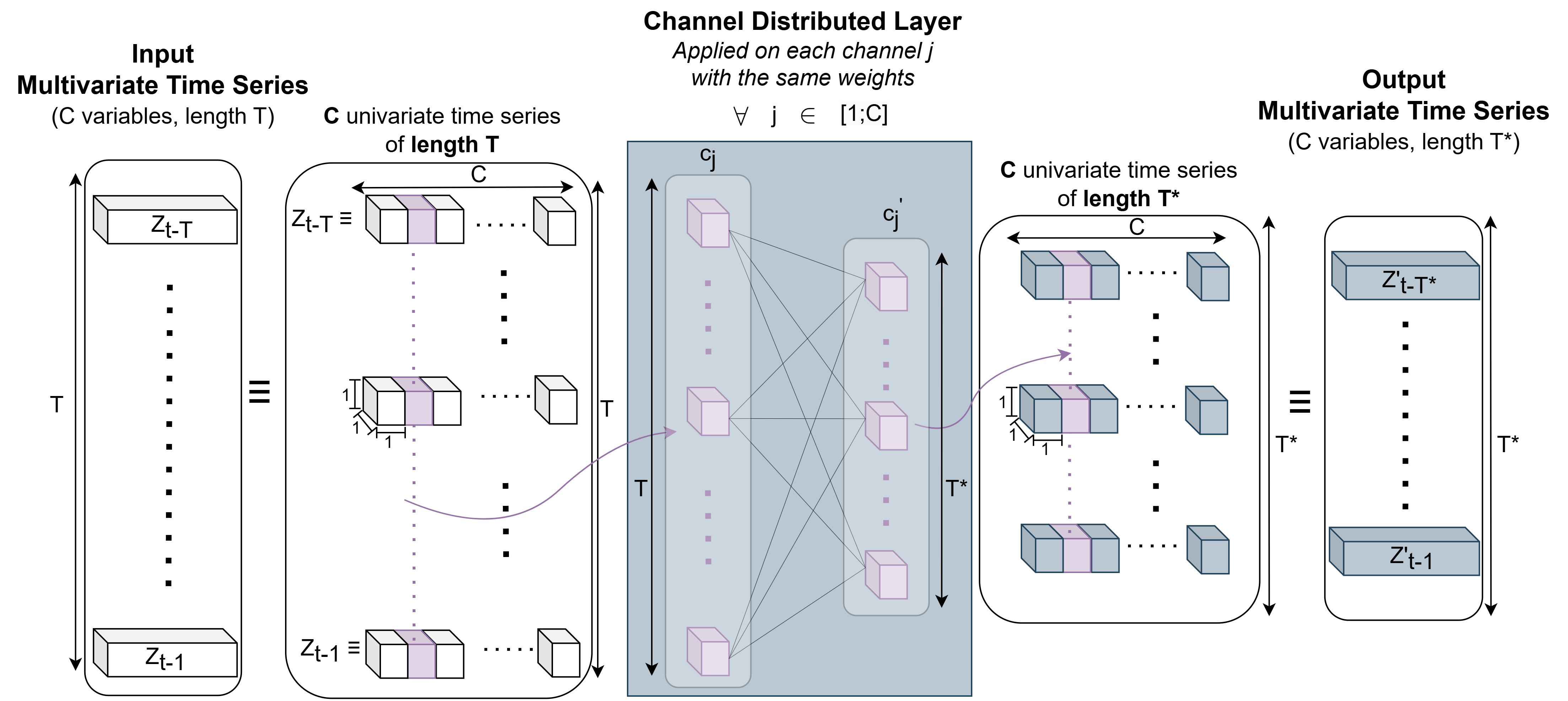}
        \caption{}
        \label{fig:cd_layer}
    \end{subfigure}
\caption{\textbf{a)} Time Distributed (TD) Layer with a fully connected cell \textbf{b)} Channel Distributed (CD) Layer with a fully connected cell.}
\label{fig:td_cd_layers}
\end{figure*}

While analyzing the TD layers, a question caught our attention:
\textit{Why not apply the behavior of TD layer to channels (i.e. variables) instead of time?} 
This implies processing each channel individually and performing computations on the temporal dimension. In this fashion, the channel-wise information is preserved, while the sequential (temporal) information is processed with the same cell for each channel.
This brought us to develop the Channel Distributed (CD) cell-based layers that, as explained in the following, have been adopted in the development of the UnPWaveNet architecture.\\
Instead of looking at the multivariate time series $Z$ as a series of $T$ time-indexed vectors each with $C$ element, it is possible to interpret it as a set of $C$ univariate time series $Z = \{c_j \in \mathbb{R}^T |\ j \in [1; C]$\}.
Figure~\ref{fig:cd_layer} shows how a CD layer with a fully connected cell works. More specifically, in Figure~\ref{fig:cd_layer} the CD layer acts channel by channel, transforming each univariate time series $c_j \in \mathbb{R}^T$ into a new series $c_j' \in \mathbb{R}^{T^*}$ $\forall\ j\in[1;C]$.
If the number of neurons $T^*$ in the cell is less than the number of elements in the input univariate series (i.e. $T < T^*$), the CD layer compresses the time dimension of the input sequence, leaving untouched the channel dimension which still contains $C$ variables.
Reversely, if $T > T*$ the CD layer will expand the time dimension.
This means that the CD layer squeezes or dilates the time dimension of all channels with the same cell; that is exactly what the TD layer does but acting instead on the time dimension.
As for the TD layer, the cell could take any form. However, in our application we found the fully connected cell to produce already satisfying results in the CD layer of the UnPWaveNet.

\subsection{WaveNet \& UnPWaveNet}
\label{sec:methods_wavenet}

\begin{figure*}[t!]
\centering

    \begin{subfigure}[c]{0.65\textwidth}
        \centering
        \includegraphics[width=\textwidth]{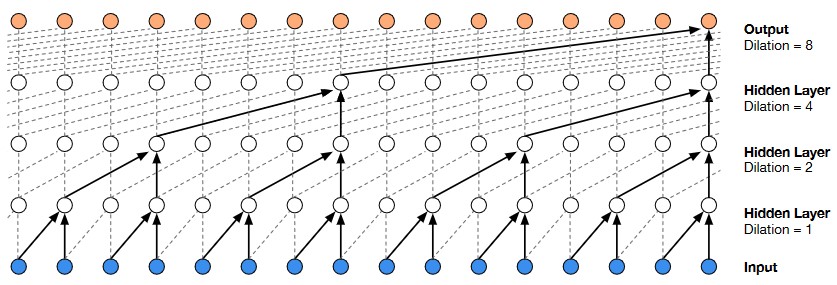}
	\caption{}
	\label{fig:causal_dilated}
    \end{subfigure}
    %\hfill
    \begin{subfigure}[c]{0.65\textwidth}
        \centering
        \includegraphics[width=\textwidth]{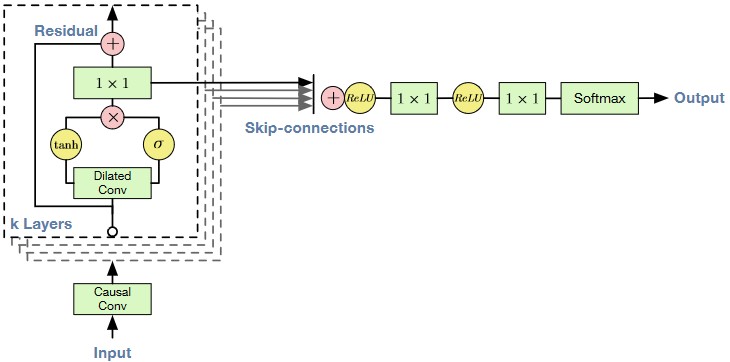}
	\caption{}
	\label{fig:wavenet}
    \end{subfigure}
\caption{\textbf{a)} Dilated causal convolution as implemented in the original WaveNet, Figure 3 in \cite{OordWaveNet2016}. In the picture, each convolution has a kernel size of 2, and the dilation increases exponentially. \textbf{b)} Original Wavenet architecture, Figure 4 in \cite{OordWaveNet2016}.}
\label{fig:original_wavenet_all}
\end{figure*}

\begin{figure*}[t!]
\centering

    \begin{subfigure}[c]{0.7\textwidth}
        \centering
        \includegraphics[width=\textwidth]{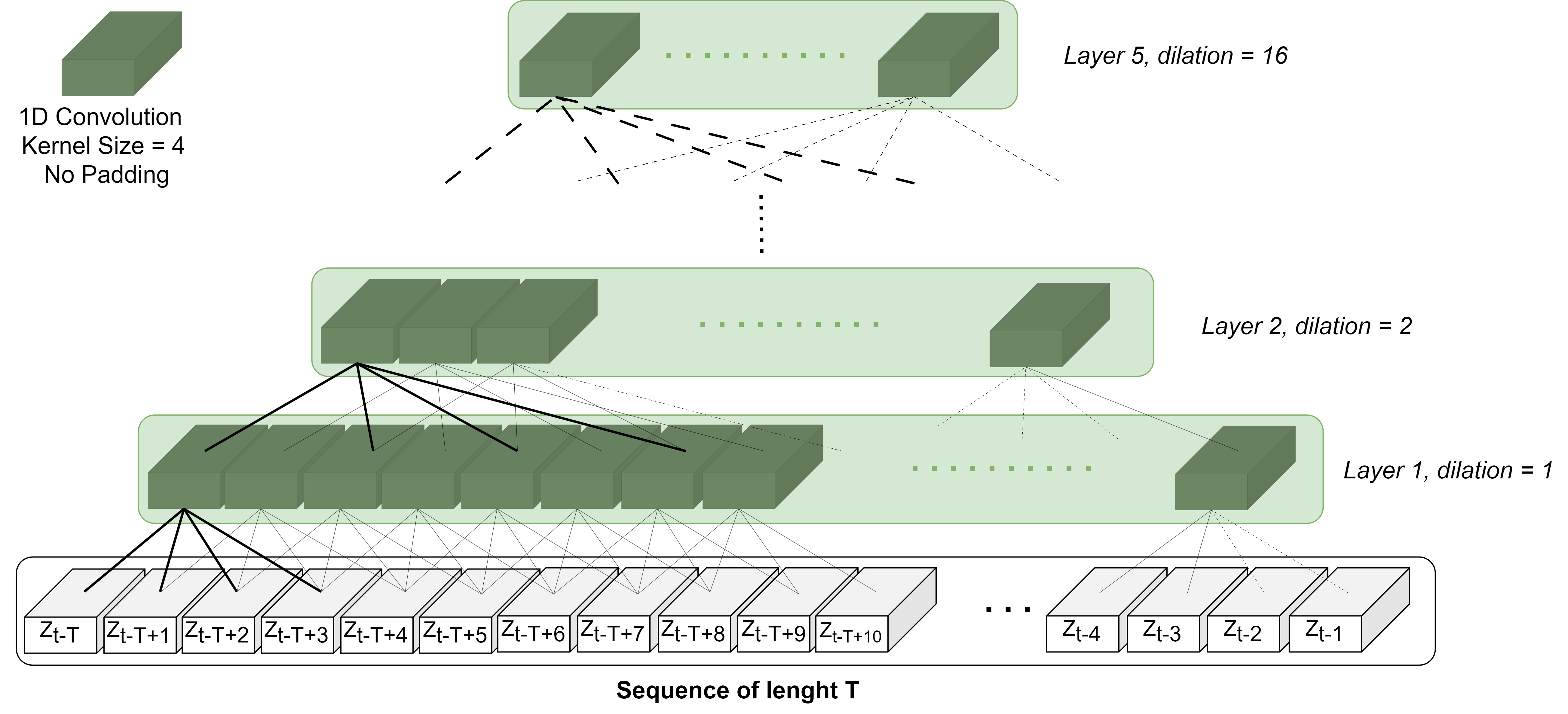}
	\caption{}
	\label{fig:unp_dilated_conv}
    \end{subfigure}
    %\hfill
    \begin{subfigure}[c]{0.7\textwidth}
        \centering
        \includegraphics[width=\textwidth]{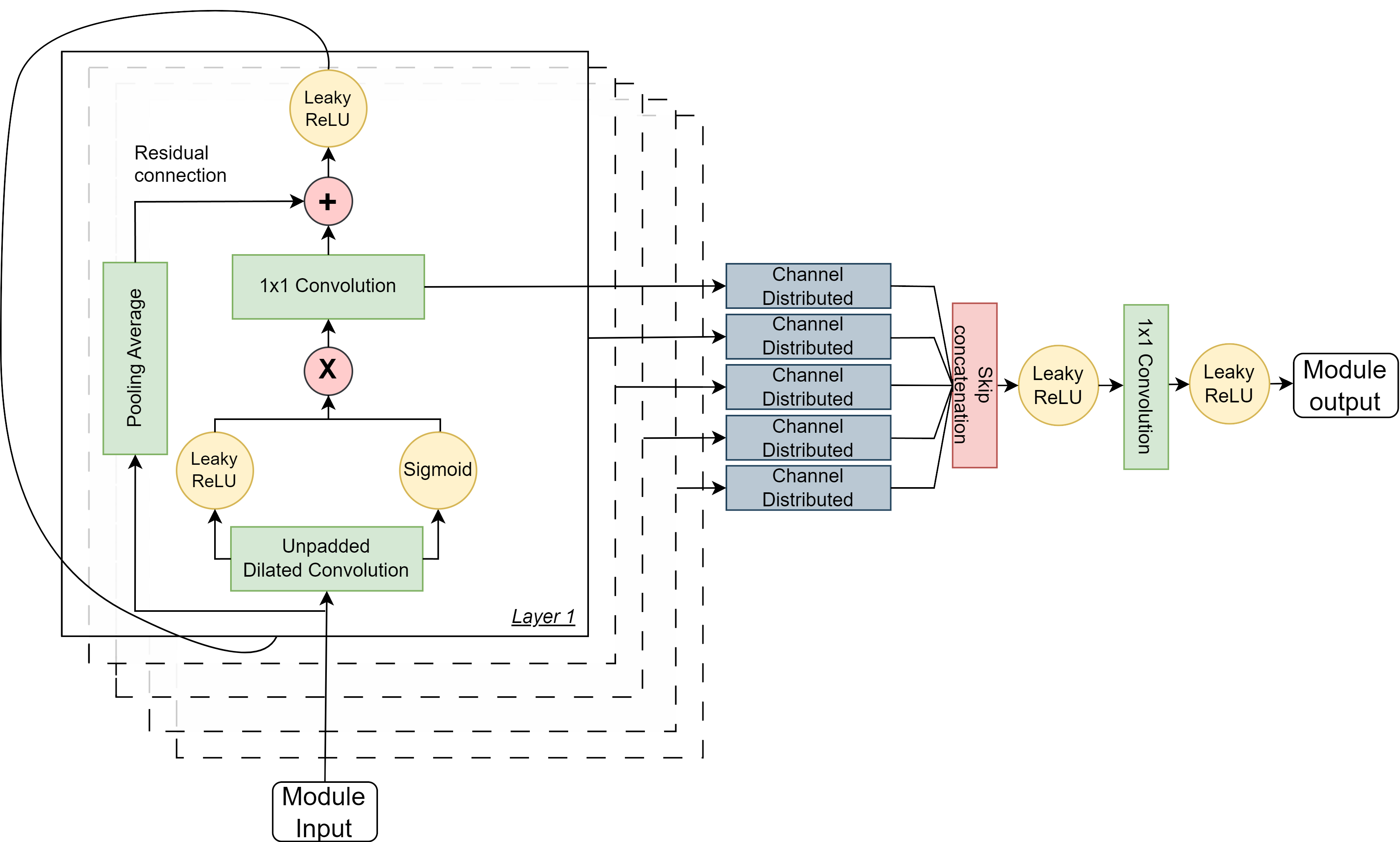}
	\caption{}
	\label{fig:unpwavenet_arch}
    \end{subfigure}
\caption{\textbf{a)} Unpadded Dilated Convolution. In the picture, each convolution has a kernel size of 4, and the dilation increases exponentially. \textbf{b)} UnPWaveNet architecture.}
\label{fig:UnPWaveNet}
\end{figure*}

In the last year, many studies have tried to develop new convolutional models for temporal-sequential data tasks to compete with RNN\cite{LeaTemporal2016,GehringConvolutional2017,BaiEmpirical2018,BaiTrellis2019,IsmailFawazInceptionTime2020,LiModeling2022}.
Most of these works are based on dilated convolution, which enables the exponential expansion of the receptive field of the network over the input sequence (i.e. look far away in the past of the input series) \cite{YuMultiScale2016}.
WaveNet \cite{OordWaveNet2016} is exactly based on this concept, and it also integrates a causal constraint using causal padding.
The causal padding makes every element of the output sequence to depend only on current and past input data, and not on the future. Furthermore, it forces the output to have the same length as the input sequence. All this is achieved by sliding the convolution operations from right (more recent values) to left (older values) and adding zeros to the left of the input\footnote{For more details on the dilated convolution and causal passing look at \cite{YuMultiScale2016,LeaTemporal2016,OordWaveNet2016,BaiEmpirical2018,BorovykhConditional2018}}.
In \cite{OordWaveNet2016}, this processing is named dilated causal convolution, shown in Figure~\ref{fig:causal_dilated}. 
Even if the WaveNet model was designed for audio generation, thanks to its ability to handle temporal data, it has been applied with remarkable results also to other tasks, among which financial data \cite{BorovykhConditional2018} and hydrology \cite{ChenWaveNetbased2023}.
Each layer of the WaveNet (Figure~\ref{fig:wavenet}) consists of a dilated causal convolution, a gated activation unit \cite{OordConditional2016}\footnote{Gate activation units are also employed in the LSTM cell, in which are named \textit{gates}. Other works employed and improved this type of activation \cite{DauphinLanguage2017} obtaining better results than using the classic ReLU.}, and a 1x1 convolution\footnote{The 1x1 convolution is convolution with a kernel of dimension 1 which acts as a bottleneck squeezing the channel dimension \cite{HeDeep2015}.  This is equivalent to a TD fully connected layer with as many neurons as the number of filters of the convolution.}. 
What makes WaveNet very flexible are the residual connections over the dilated causal convolutions and the skip connections which concatenate the result of every 1x1 convolution letting the gradient flow easily over the network. \\
Even if the WaveNet and other convolutional-based networks have brought astonishing results, they have been usually applied for many-to-many tasks, and in general to predict sequences of the same length of the input.
We have restructured the WaveNet architecture to predict output sequences completely shifted in the future and shorter than the input sequences. 
In this case, the causal constraint is no longer needed, because the output is completely in the future and, thus each output element should depend on every input element. 
Consequently, in such a case, it is possible to drop the causal padding implemented in the WaveNet and let the temporal dimension be squeezed layer after layer.
Figure~\ref{fig:unp_dilated_conv} represents dilated convolution without the causal padding, here named as unpadded dilated convolution, whose receptive field $r_l$ for a layer $l$ with a filter of dimension $K$ is defined by Equation~\ref{eq:receptive_field_dconv}. It is worth noting that without the causal constraint, the first element of the output sequence of a layer $l$ has a receptive field $r_l$ which covers the first $r_l$ elements of the input sequence, while the last element of the output sequence has a receptive field which covers the last $r_l$ elements of the input sequence.
Instead, when the causal constraint is enforced, every element of the output sequence of a layer $l$ has a receptive field that covers up to $r_l$ first (i.e. past) element of the input sequence.

\begin{equation}
r_l = 1 + (K - 1) \sum_{i = 1}^{l} 2^{i-1}  
\label{eq:receptive_field_dconv}
\end{equation}

Two problems arise when removing the causal padding constraint from the initial network architecture: it is a) no longer possible to add residual connections; and b) no longer possible to concatenate skip connections. 
The cause for both problems is the different dimensions of the input and output sequences of each layer, which without the padding are no longer maintained over the network.
To solve the problem a), we apply a pooling average layer to meet the dimension of the 1x1 convolution.
While to solve problem b) we have adopted our proposed Channel Distributed layer with a fully connected cell, which transforms the output sequences from 1x1 convolutions into new sequences with the same length equal to the length of the requested output.
Figure~\ref{fig:unpwavenet_arch} shows the architecture of our proposed version of the WaveNet, here renamed UnPWaveNet (unpadded WaveNet).
With respect to the original implementation (Figure~\ref{fig:wavenet}), the UnPWaveNet uses only one 1x1 convolution layer instead of two after the skip concatenation.
This is because the CD layers already apply some transformation to the skip connections, and thus it was sufficient to achieve satisfactory results saving parameters. \\
In our specific case study, the output sequence is far shorter than the input, so that it is only a scalar, thus a many-to-one scenario. However, the UnPWaveNet architecture could be applied in any many-to-many case in which the output sequence is completely shifted in the future and shorter than the input one.

\subsection{Proposed models}
\label{sec:methods_models}

\subsubsection{TDC module}
\label{sec:methods_models_TDCmodule}

\begin{figure*}[t]
	\centering
	\includegraphics[width=.85\textwidth]{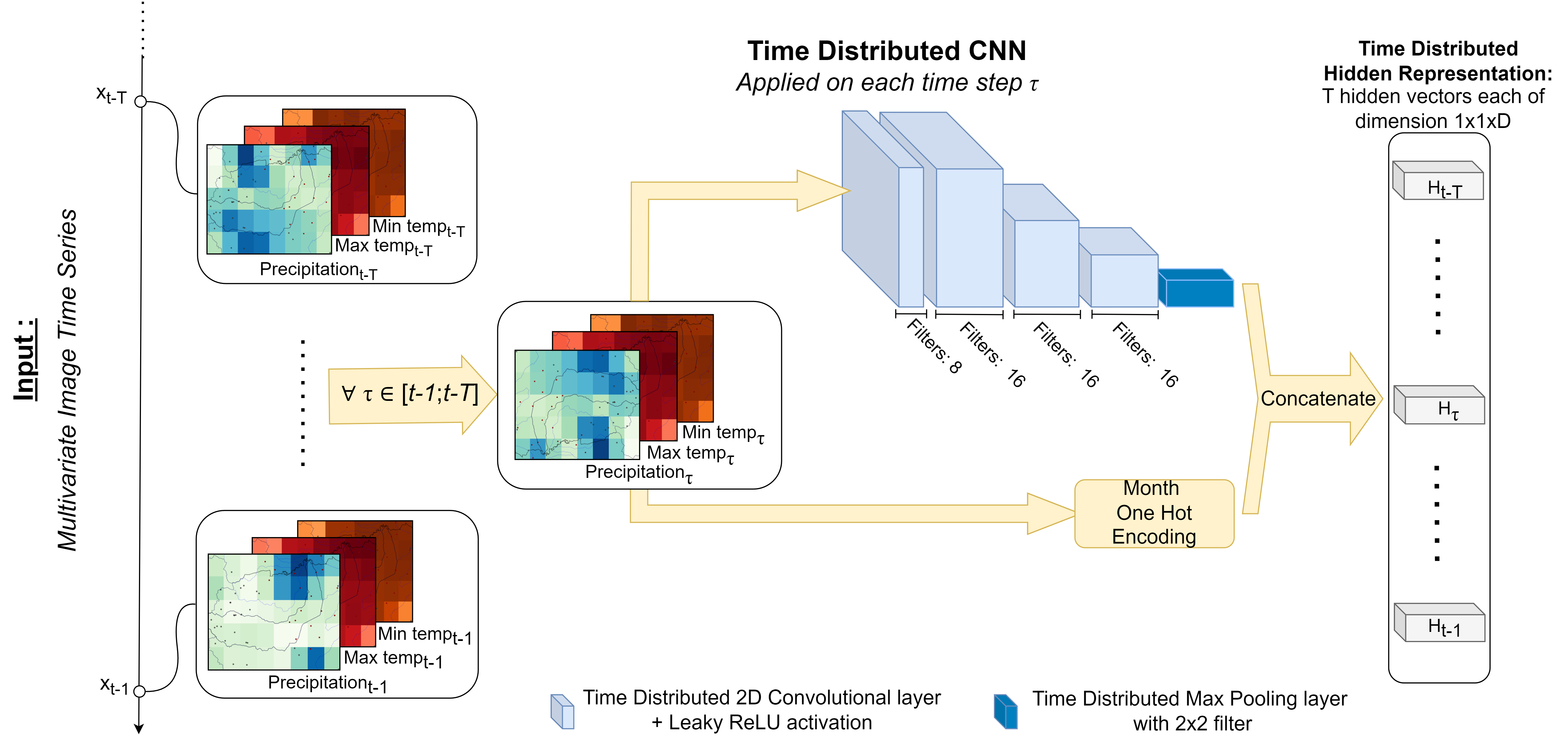}
	\caption{Time Distributed CNN (TDC) module.}
	\label{fig:tdc_module}
\end{figure*}

The TDC module is responsible for learning a vectorial representation of the images available at each time step.
Thus, it converts the input image time series (i.e. video) into a classical multivariate time series. 
Figure~\ref{fig:tdc_module} depicts the architecture of the TDC module. It is made up of a TD CNN which takes the image with $P$ channels available at every time $\tau$ and feeds it into a 4-layer CNN with filters of size 2 and leaky-ReLU activation function. These 4 layers reduce the spatial dimension and increase the channels (see the number of filters shown in Figure~\ref{fig:tdc_module}). The last max pooling layer is then responsible for squeezing the spatial extent and outputs a vector with 16 elements.\\
Along with the convolutional operations, the one hot encoding (OHE) of the corresponding month of $\tau$ is computed using 11-dimensional vector\footnote{Encoding 12 exclusive categories into a 12-dimensional binary vector would have brought to linear dependent features, i.e. perfect collinearity.}; this lets the network to take account of seasonality behaviors. \\
The output of the TDC module is then, $\forall \tau \in [t-1,t-T]$, the concatenation of the OHE, and the max pooling output. This is a multivariate time series with D variables, here named Time Distributed Hidden Representation.
In our case study, $D$ is equal to 27 (16 max pooling dimension plus 11 OHE); and $P$ is equal to 3, i.e. the three weather variables employed: total precipitation, maximum temperature, and minimum temperature.
The TDC module is employed in both two models, TDC-LSTM and TDC-UnPWaveNet, with the same structure and hyperparameters. 

\subsubsection{TDC-LSTM model}
\label{sec:methods_models_TDCLSTM}

\begin{figure*}[t]
	\centering
	\includegraphics[width=0.85\textwidth]{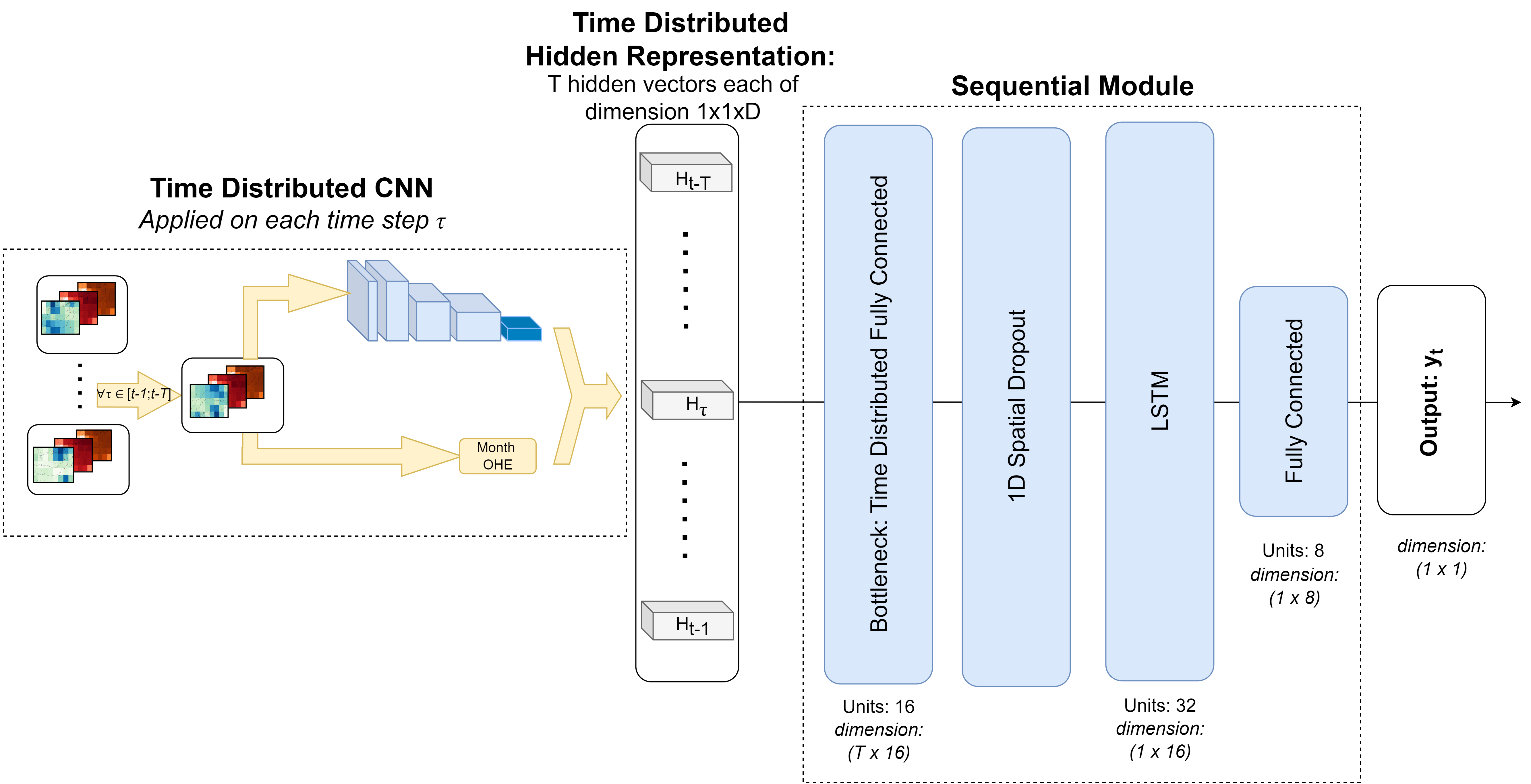}
	\caption{TDC-LSTM architecture. Under each layer output dimensions are reported.}
	\label{fig:tdc_lstm}
\end{figure*}

The TDC-LSTM model uses the TDC module and then a Sequential Module as depicted in Figure~\ref{fig:tdc_lstm}.
In detail, a first bottleneck layer made of a TD fully connected followed by a Leaky-ReLU activation reduces the channel dimension to 16. This decreases the number of parameters in the subsequent layers, and thus it helps also in mitigating overfitting \cite{HeDeep2015, IsmailFawazInceptionTime2020}.
Then, a 1D spatial dropout with probability $0.15$ is employed as a regularization technique. Instead of the classical dropout, the spatial dropout zero-out entire channels and not single element inside channels; this is done to face correlation issues between consecutive elements in a channel \cite{TompsonEfficient2015}. A single LSTM layer with 32 units is then adopted to model the temporal relations, leaky-ReLU, and sigmoid are employed as activation functions for the gates inside the LSTM cells. Leaky-ReLU has proved to be more effective than tanh in this task providing better results.
A fully connected layer with 8 neurons and leaky-ReLU is then used to reduce the dimensionality of the 32-dimensional output of the LSTM. The last output layer computes an affine transformation and outputs the water table depth.
The TDC-LSTM model has 9705 parameters in total.
% 6272 lstm

\subsubsection{TDC-UnPWaveNet model}
\label{sec:methods_models_TDCUnPWaveNet}

\begin{figure*}[t]
	\centering
	\includegraphics[width=0.85\textwidth]{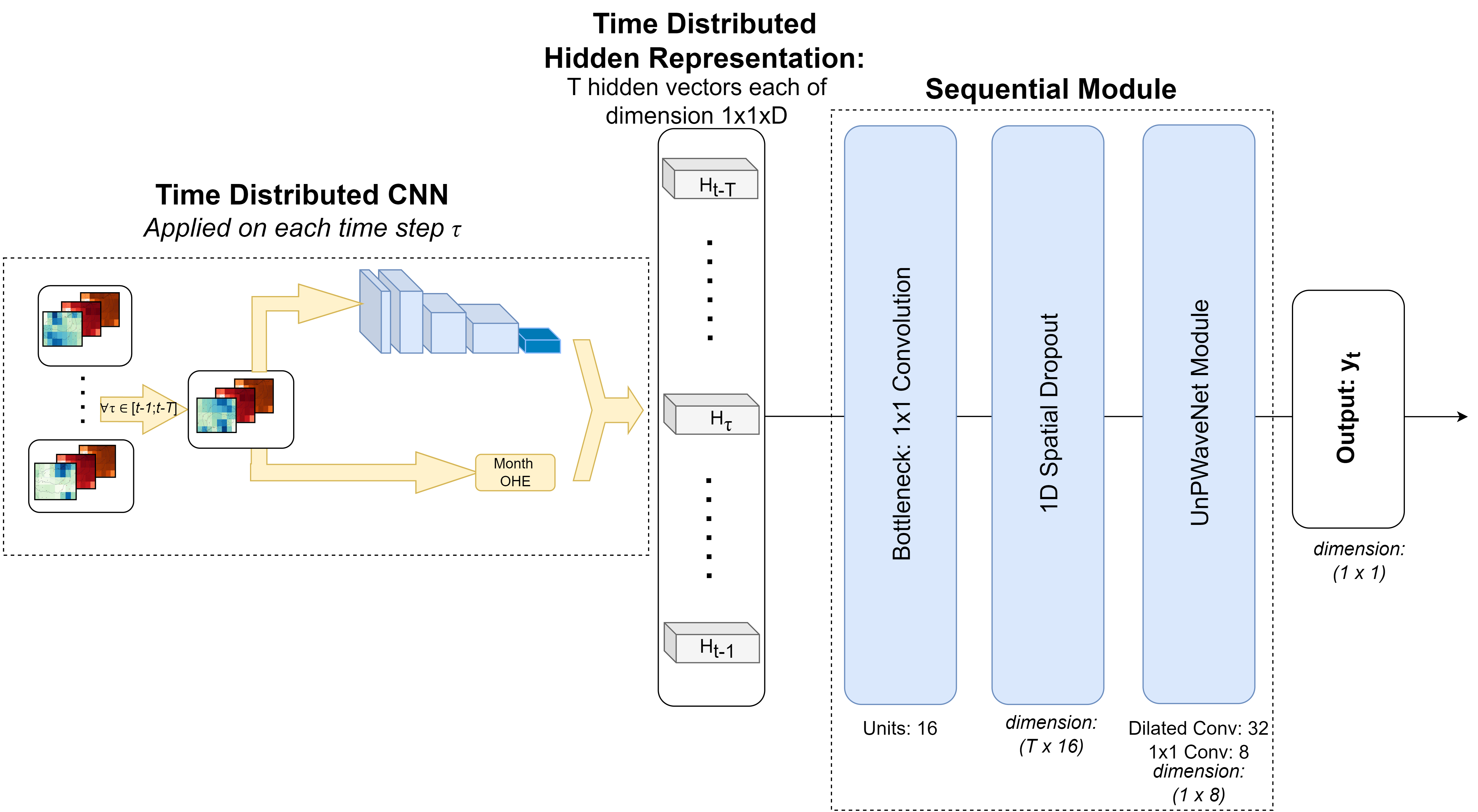}
	\caption{TDC-UnPWaveNet architecture. Under each layer output dimensions are reported.}
	\label{fig:tdc_unpwavenet}
\end{figure*}

The structure of the TDC-UnPWaveNet is very similar to the TDC-LSTM. The TDC module is still the same, however, the Sequential Module adopts the UnPWaveNet for learning temporal relations.
Figure~\ref{fig:tdc_unpwavenet} depicts the architecture of the TDC-UnPWaveNet models.
As for the TDC-LSTM, the first layer is still a bottleneck layer which reduces the channel dimension to 16.
The bottleneck layer here is implemented as a 1x1 convolution, but as already stated in Section~\ref{sec:methods_wavenet}, it is equivalent to a TD fully-connected layer with the number of neurons equal to the number of filters of the 1x1 convolution.
Then, the spatial dropout with probability $0.15$ is applied, and its output is fed into the UnPWaveNet module (Figure~\ref{fig:unpwavenet_arch}) which outputs an 8-dimensional vector that is fed into the last output layer identical to the TDC-LSTM model. 
The UnPWaveNet module is implemented using 5 layers of unpadded dilated convolution with 32 filters of size 4, and 1x1 convolution with 8 filters. The dilation is set $2^{l-1}$ for each layer $l$; in this way, the last (5th) layer has a dilation of 16.
The TDC-UnPWaveNet model has 17915 parameters in total.
% 14746.

\subsection{Implementation details}

\subsubsection{Preprocessing}
\label{sec:methods_preprocessing}

\begin{figure*}[t]
	\centering
	\includegraphics[width=.85\textwidth]{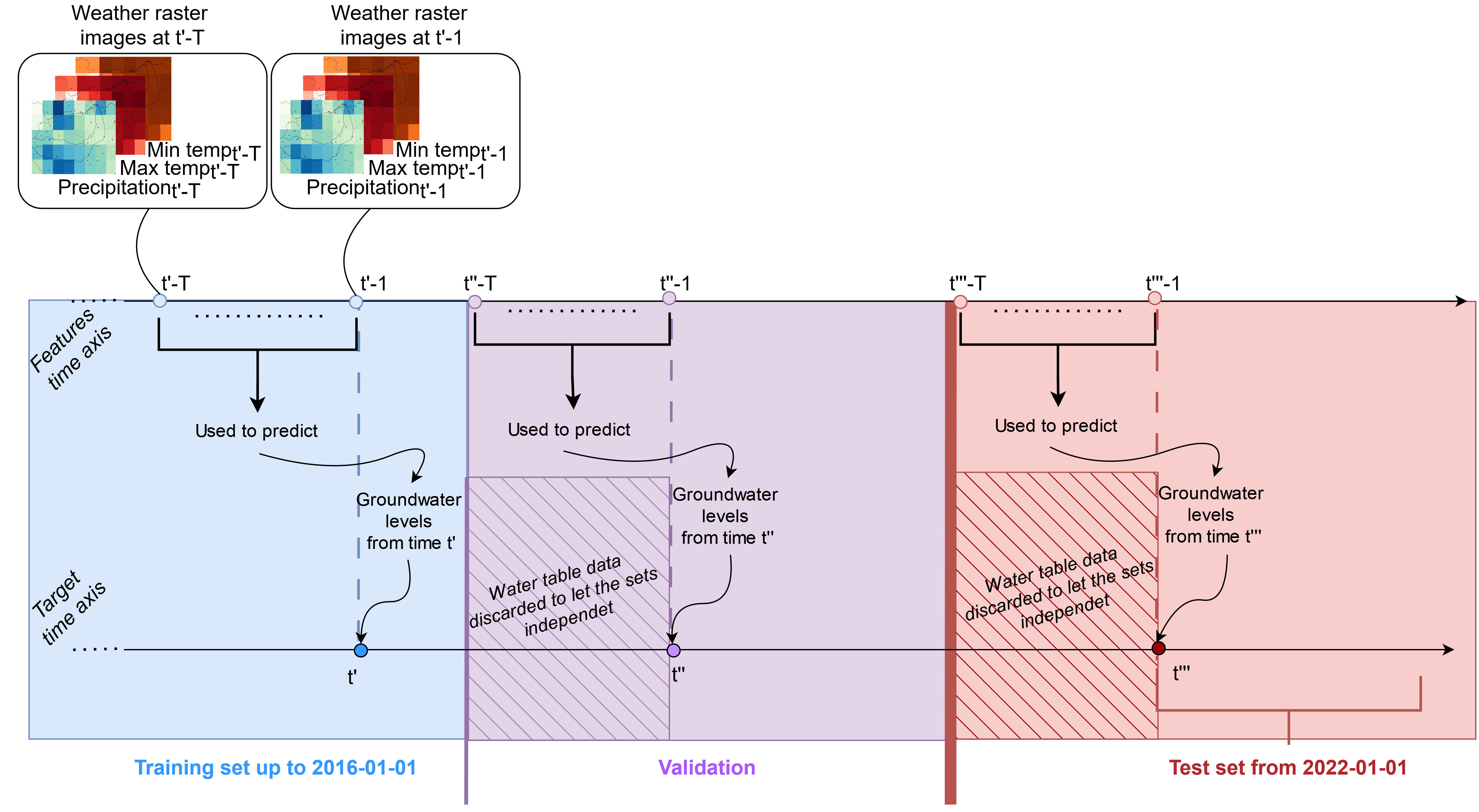}
	\caption{Training, validation, and test definition.}
	\label{fig:set_split}
\end{figure*}

The raw weather raster images covered the entire Piedmont region, to focus on the catchment all the images were clipped on the ROI maintaining a squared shape which is easier to handle with CNNs. A box with a lower-left corner in coordinate (6.90°E;44.35°N) and higher-right in (7.79°E;44.84°N) was adopted to clip the images.
Concerning the temporal resolution, we set a weekly time step for the predictions, and then both the target and features were aggregated computing weekly averages. \\
In a time series task in which lagged features are employed, inserting gaps between the training, validation, and test sets is common to prevent data leakage and performance overestimation.
For example, in the case of an autoregressive model that uses lags up to $t-T$ as features to predict the target at time $t$, it is usual to discard the $T$ time steps between sets. 
In the present case study, even if the models do not use autoregressive terms but only exogenous (weather) variables, we think that a gap is still needed because the last observations in a set use almost the same predictor as the first observation in the next set. 
Then, especially in cases where $T$ is large, this affects the independence of consecutive sets, and then, the unbiasedness of the performance metrics. More formally it is possible to speak about leakage from training examples as described in \cite{KaufmanLeakage2012}. 
Many related studies as \cite{WunschGroundwater2021,AndersonEvaluation2022} did not mention this issue, and built consecutive and overlapping sets. 
Here, instead, in a more precautionary fashion, a gap of $T$ time step is considered for defining sets. In other words, we have constrained that features used in a set can not be used as predictors also in a subsequent set.\\
A problem related to the introduction of gaps of $T$ time step is the discarding of $T$ observations; which became a major concern if $T$ is long and the total number of observations is already low. 
Our case study fits partially to this worst case because we adopted a very long $T$, however, even if the total observation for each sensor is not very large (See Table~\ref{tab:wt_statistics}), the proposed models have yielded very satisfactory results even with the introduction of the gaps between consecutive sets.
To introduce the gaps, we have attempted to exploit most of the already missing periods in our data.
Furthermore, we have defined a splitting rule trying to set test periods as much overlapping as possible for all the three series.
In detail, for each sensor, we have considered the training set as the water table data available up to 2016-01-01; the test as the data from 2022-01-01 onward; and validation as the remaining data between training ending time and test beginning time (see Figure~\ref{fig:set_split} for a better understanding).\\
Normalizing the data is an effective practice in data science because it eases the learning process, especially when different exogenous variables are employed. For this reason, all the data have been normalized by computing z-scores using the corresponding means and standard deviations of the training set: $z = \frac{x - \overline{x_{training}}}{\sigma_{x_{training}}} $.

\subsubsection{Hyperparameters and training}
\label{sec:methods_hyperpar}

Both the TDC-LSTM and TDC-UnPWaveNet have been trained with stochastic gradient descent with momentum and Nesterov \cite{GoodfellowDeep2016} using Mean Squared Error (MSE) as a loss function. The number of epochs has been fixed to 80 and the batch size to 8.
A clipnorm value of $1.0$ has been adopted to face the problem of exploding gradient.
L2 regularization has been used to tackle overfitting and increase the generalization ability of the models. In table~\ref{tab:hyperparams} the learning rates and L2 regularization are shown for each local model. All these hyperparameters have been found by a manual grid search strategy.\\
To take into account the uncertainty of the random initialization of the weight we have initialized and trained 10 times each local model independently. We have considered the ensemble mean as the final prediction for each local model.\\
All the experiment was performed in Python using the Colab environment and its freely available hardware. To develop DL models Tensorflow and Keras 2.15.0 were used.

\subsubsection{Evaluation metrics}
Many performance metrics have been computed following Equations~\ref{eq:rmse},~\ref{eq:nrmse},~\ref{eq:bias},~\ref{eq:nbias},~\ref{eq:mape},~\ref{eq:rho},~\ref{eq:nse}, and~\ref{eq:kge}. 
In these equations $\overline{y}$, $y_{min}$, $y_{max}$ represent respectively the target mean, minimum, and maximum computed over the training set. 
NSE and KGE have been frequently adopted in hydrological modeling studies. More in detail, NSE shows how a model performs with respect to a naive estimator, which is usually the mean ($\overline{y}$). The NSE has an intrinsic benchmark set at 0, which is when the model performs as well as the naive estimator. 
The KGE \cite{GuptaDecomposition2009} is a different concept, it is the Euclidean distance between the vector defined by metrics $\rho$ (as in Equation~\ref{eq:rho}),$\alpha$ and, $\beta$ and the vector with the best achievable metrics ($\rho = 1$, $\alpha = 1$ and, $\beta = 1$). 
In practice, the KGE is a measure that takes into account more aspects of the prediction, i.e. the Pearson correlation, the bias, and the variance.
The KGE has no intrinsic benchmark as the NSE and, as pointed out in \cite{KnobenTechnical2019}, these two metrics are not directly comparable.
In \cite{KnobenTechnical2019} authors stated that if the benchmark is set as the NSE. i.e. a fixed mean estimation, then the cut-off point of the KGE is $-0.41$, after which the model performs better than the naive estimator. 
For both NSE and KGE the maximum value is 1 and the higher the values the better a model is performing.

\begin{table*}[t]
\centering
\caption{Hyperparameters}
\begin{tabularx}{\textwidth}{YY|YY} %\tblwidth
\toprule
\textbf{Sensor} & \textbf{Model}   & \textbf{Learning Rate} & \textbf{L2 regularization}\\ 
\hline\hline
\multirow{2}{*}{\shortstack{\textbf{Vottignasco} \\ \textbf{00425010001}}} %
& \textit{TD CNN + LSTM}   &  0.001 &  0.0025 \\ 
& \textit{TD CNN + UnPWaveNet} &  0.0025 &  0.0075 \\ \hline
\multirow{2}{*}{\shortstack{\textbf{Savigliano} \\ \textbf{00421510001}}}  
& \textit{TD CNN + LSTM}    & 0.001   & 0.00075     \\ 
& \textit{TD CNN + UnPWaveNet} & 0.001  & 0.0075\\ \hline
\multirow{2}{*}{\shortstack{\textbf{Racconigi} \\ \textbf{00417910001}}}  
& \textit{TD CNN + LSTM}    & 0.001  & 0.0005     \\ 
& \textit{TD CNN + UnPWaveNet}  & 0.001  & 0.0075\\
\bottomrule
\end{tabularx}
\label{tab:hyperparams}
\end{table*}

\begin{equation}
RMSE = \sqrt{\frac{1}{N}\sum_{i}^{N} ( \hat{y_i} - y_i )^2} 
\label{eq:rmse}
\end{equation}

\begin{equation}
NRMSE = \frac{\sqrt{\frac{1}{N}\sum_{i}^{N} ( \hat{y_i} - y_i )^2}}{y_{max} - y_{min} } 
\label{eq:nrmse}
\end{equation}

\begin{equation}
BIAS = \frac{1}{N}\sum_{i}^{N} (\hat{y_i} - y_i)
\label{eq:bias}
\end{equation}

\begin{equation}
NBIAS = \frac{\frac{1}{N}\sum_{i}^{N} (\hat{y_i} - y_i)}{y_{max} - y_{min} } 
\label{eq:nbias}
\end{equation}

\begin{equation}
MAPE = \frac{1}{N}\sum_{i}^{N} \frac{|\hat{y_i} - y_i|}{y_i} 
\label{eq:mape}
\end{equation}

\begin{equation}
\rho = \frac{\sum_{i}^{N}(\hat{y_i}-\overline{\hat{y}})(y_i-\overline{y})}{\sqrt{\sum_{i}^{N}(\hat{y_i}-\overline{\hat{y}})^2 \sum_{i}^{N}(y_i-\overline{y})^2}} 
\label{eq:rho}
\end{equation}

\begin{equation}
NSE = 1 - \frac{\sum_{i}^{N} ( \hat{y_i} - y_i )^2}{\sum_{i}^{N} ( y_i - \overline{y}  )^2}
\label{eq:nse}
\end{equation}

\begin{equation}
    \begin{gathered}
    KGE = 1 - \sqrt{(\rho-1)^2 + (\alpha - 1 )^2 + (\beta - 1)^2 } \\
    \alpha = \frac{\sigma_{\hat{y}} }{\sigma_y};
    \beta = \frac{\mu_{\hat{y}} }{\mu_y}
    \end{gathered} 
   \label{eq:kge}
\end{equation}

\section{Results}\label{sec:results}

\begin{table*}[t]
\centering

\caption{Performance metrics on test sets computed on the ensemble mean predictions. The last row reports the means and the standard deviations in parenthesis over the three sensors.}
\begin{tabularx}{\textwidth}{@{\hspace{0.1cm}}cc@{\hspace{0.1cm}}|@{\hspace{0.1cm}}c@{\hspace{0.15cm}}YYYY@{\hspace{0.2cm}}c@{\hspace{0.2cm}}Yc@{\hspace{0.1cm}}} 
\toprule
\textbf{Sensor} & \textbf{Model} & \textbf{RMSE[m]} & \textbf{NRMSE} & \textbf{BIAS[m] } & \textbf{NBIAS} & \textbf{MAPE} & \textbf{$\rho$} & \textbf{NSE} & \textbf{KGE}\\ 
\hline\hline
\multirow{2}{*}{\shortstack{\textbf{Vottignasco} \\\textbf{00425010001}}}
& \textit{TDC-LSTM} & 0.37 & 0.09 & -0.21 & -0.05	& 0.04 & 0.94 & 0.93 & 0.36\\ 
& \textit{TDC-UnPWaveNet} & 0.38 & 0.09 & -0.28 & -0.06 &	0.05 &	0.96 & 0.93 &	0.39\\ \hline

\multirow{2}{*}{\shortstack{\textbf{Savigliano} \\\textbf{00421510001}}}  
  & \textit{TDC-LSTM} & 0.11 & 0.08 & -0.10 & -0.07 & 0.02 & 0.97 & 0.90 & 0.52\\ 
  & \textit{TDC-UnPWaveNet} & 0.05 & 0.04 & $<0.01$ & $<0.01$ & 0.01 & 0.96 & 0.98 & 0.51\\ \hline

\multirow{2}{*}{\shortstack{\textbf{Racconigi} \\\textbf{00417910001}}}  
  & \textit{TDC-LSTM} & 0.34 & 0.07 & -0.22 & -0.04 & 0.05 & 0.89 & 0.90 & 0.37\\ 
  & \textit{TDC-UnPWaveNet} & 0.49 & 0.10 & -0.46 & -0.09 & 0.08 & 0.95 & 0.79 & 0.40\\

\hline\hline
\multirow{4}{*}{\shortstack{\textbf{Mean} \\ ($\sigma$) \cr \cr \cr \cr}}  
  & \textit{TDC-LSTM} & 0.27 & 0.08 & -0.18 & -0.05 & 0.04 & 0.93 & 0.91 & 0.41\\[-3pt]
  & & (0.12) & (0.01) & (0.05) & (0.01) & (0.01) & (0.03) & (0.02) & (0.07)\\ 
  & \textit{TDC-UnPWaveNet} & 0.31 & 0.08 & -0.25 & -0.05 & 0.05 & 0.96 & 0.90 & 0.43\\[-3pt]
  & & (0.19) & (0.03) & (0.19) & (0.04) & (0.03) & (0.01) & (0.08) & (0.05)\\
  
\bottomrule
\end{tabularx}
\label{tab:perf_metrics}
\end{table*}

\begin{figure*}[t!]
\centering

    \begin{subfigure}[c]{0.85\textwidth}
        \centering
        \includegraphics[width=\textwidth]{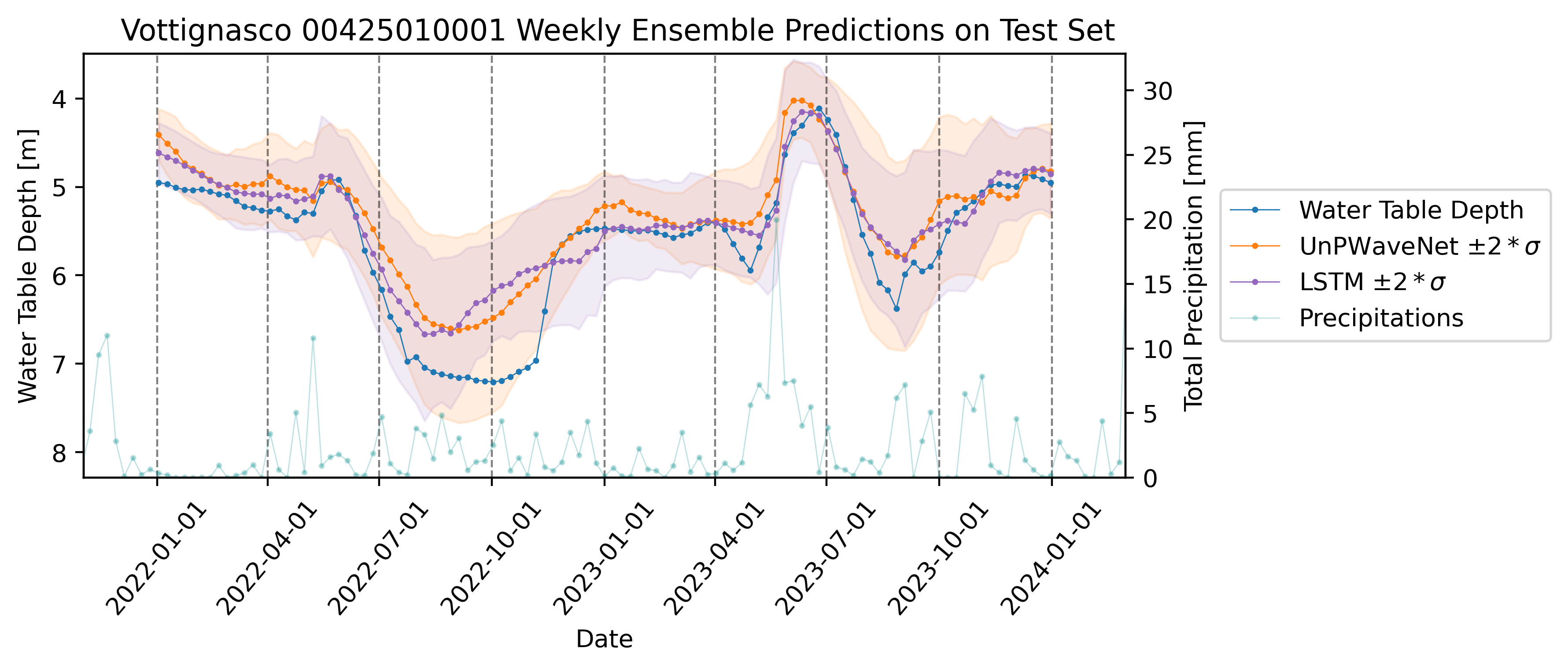}
	\caption{}
	\label{fig:vottignasco_predictions}
    \end{subfigure}
    %\hfill
    \begin{subfigure}[c]{0.85\textwidth}
        \centering
        \includegraphics[width=\textwidth]{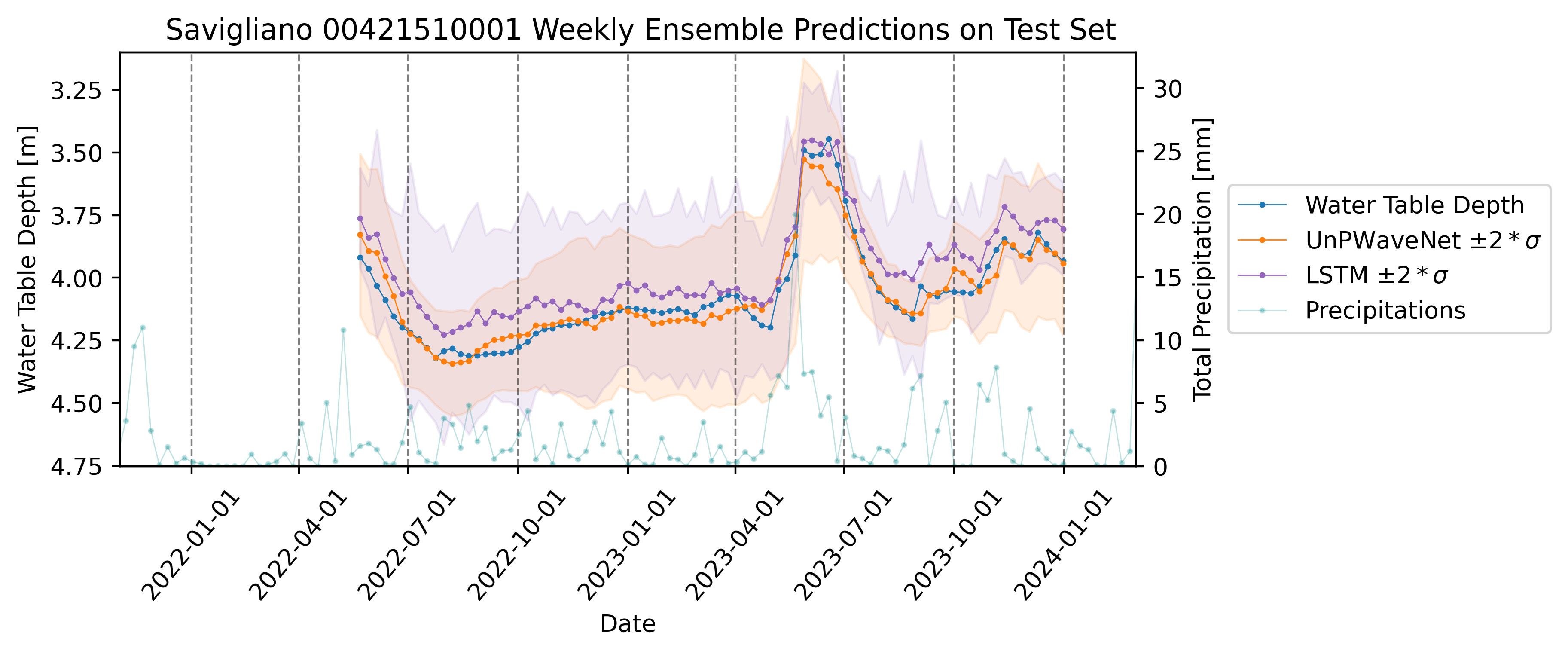}
	\caption{}
	\label{fig:savigliano_predictions}
    \end{subfigure}
    \begin{subfigure}[c]{0.85\textwidth}
        \centering
        \includegraphics[width=\textwidth]{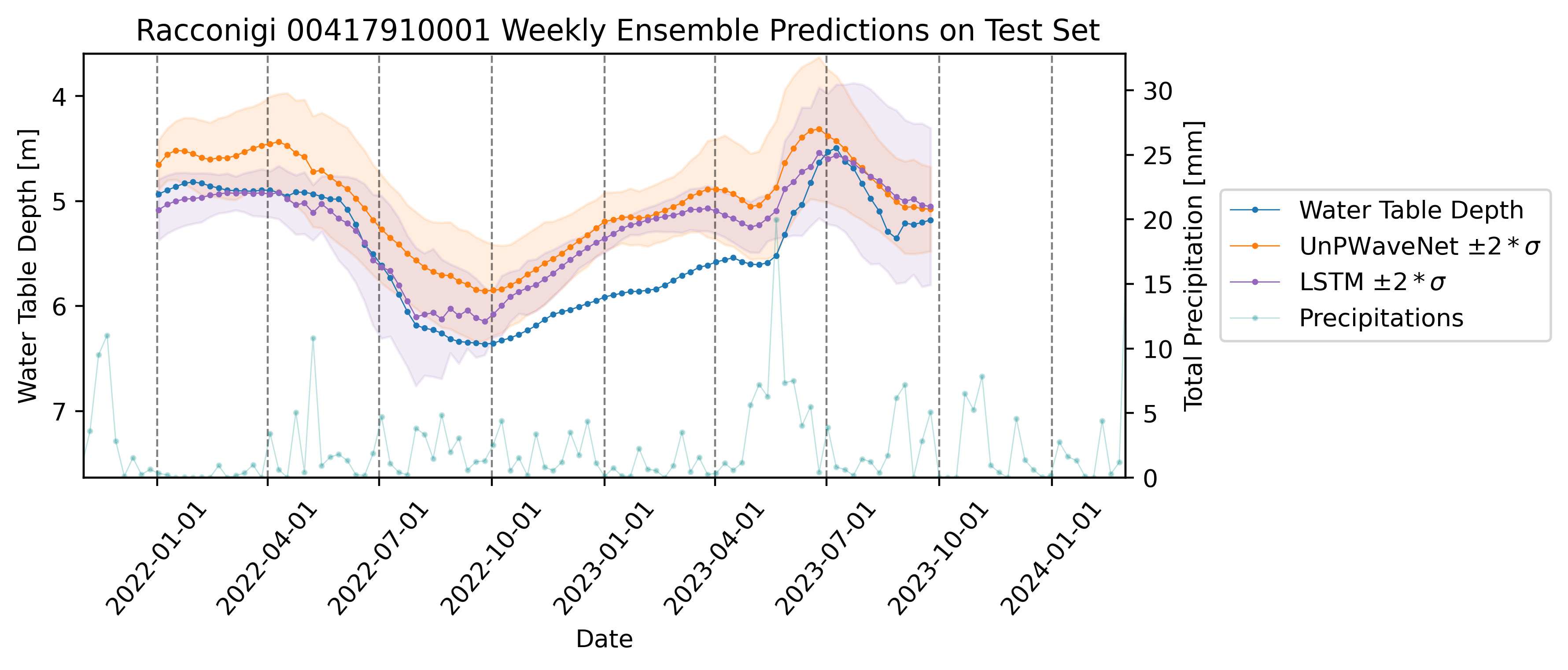}
	\caption{}
        \label{fig:racconigi_predictions}
    \end{subfigure}
\caption{Ensemble mean predictions on the test set of sensors in \textbf{a)} Vottignasco, \textbf{b)} Savigliano, \textbf{c)} Racconigi. For all the plots the shadows represent two times the standard deviation of the ensemble predictions. For shortening the legends, LSTM stands for the TDC-LSTM model, while UnPWaveNet for the TDC-UnPWaveNet model. For ease the interpretation of the water table depth y axes has been reversed.}
\label{fig:ensemble_predictions}
\end{figure*}

Table~\ref{tab:perf_metrics} shows the performance metrics computed over the test sets for the TDC-LSTM and TDC-UnPWaveNet models, the bottom row reports the mean performances over the three sensors with the corresponding standard deviations $\sigma$. Figures~\ref{fig:vottignasco_predictions}~\ref{fig:savigliano_predictions}~\ref{fig:racconigi_predictions} show the ensemble mean prediction. It is difficult to define a clear winner between the TDC-LSTM and TDC-UnPWaveNet, also because it seems that the two models have captured different aspects of the phenomenon to be modeled.
In the case of Vottignasco and Racconigi sensors, the TDC-LSTM has performed better in terms of RMSE, BIAS, MAPE, and NSE, however, the TDC-UnPWaveNet has been better for correlation and KGE. The TDC-UnPWaveNet has appeared to be more able to predict the actual temporal evolution of the ground truth, while the TDC-LSTM has been better in terms of the biasedness of predictions. 
This is very clear in Figure~\ref{fig:racconigi_predictions} in which the TDC-UnPWaveNet follows accurately the temporal evolution ($\rho = 0.95$) of the ground truth but a bias is clearly visible. 
Another example could be the drop in the Vottignasco series (Figure~\ref{fig:vottignasco_predictions}) around 2022-10-01.
Here, the TDC-UnPWaveNet has predicted correctly a more prolonged drop and more in line with the actual values, instead the TDC-LSTM has predicted a very accurate decrease of the depth, however, it incorrectly has predicted the recovery too early, probably driven by training-related memory (i.e. overfitting). In this terms, it seems that the TDC-UnPWaveNet could generalize better, at the cost of higher bias. 
For the Savigliano the TDC-UnPWaveNet has won, differences in $\rho$ and KGE are almost negligible and probably not significant from a statistical point of view. From Figure~\ref{fig:savigliano_predictions} the TDC-UnPWaveNet predictions show temporal dynamics more in line with the actual values, while the TDC-LSTM predictions appear to be far more erratic than the ground truth. Notwithstanding, for that series, the differences are very narrow and both the models have performed very well.
\\

\section{Discussion}\label{sec:discussion}

Concerning the test period, it has to be highlighted that 2022 was a very particular year in terms of weather conditions. Indeed, our ROI suffered from a severe drought in the summer which lasted until the autumn of 2022. Thus predicting the water table depth, especially in Vottignasco and Racconigi, has been a very difficult task for our models, which have to predict an uncommon drop. This has been even more difficult given the absence of an explicit autoregressive term, which would have helped the model in anchoring to the most recent actual water table depth values.

As stated in Section~\ref{sec:intro_cs_data} the choice of not using autoregressive terms in our proposed models has been guided only by a practical fact, i.e. water table depth data are updated on a semester basis.
Then, using an autoregressive term would have made our model unusable in a practical scenario to predict the next week's depth value.
Furthermore, no anthropogenic pressure proxy (e.g. human water consumption) has been fed to our model, and this is because of the lack of such data in our ROI.
Even if our models have yielded satisfactory results, they could be enhanced by the introduction of these additional inputs (as in \cite{AmarantoSemiseasonal2018,ZhangDeveloping2018,LeeUsing2019,WunschGroundwater2021}), especially to improve the performance in anomalous scenarios like the summer and autumn of 2022. Indeed, predicting such a drop could be very difficult looking only at the weather variable. For example, if the precipitations are scarce, likely, the anthropogenic pressure on the groundwater resource increases making groundwater resources decrease even more.

In \cite{WunschGroundwater2021} authors found the LSTM-based model more robust against initialization effects than CNN. We have found soft evidence of this. Indeed, in our case study, this could be true for Vottignasco and Racconigi series, in which the ensemble standard deviation of the TDC-LSTM models (shadows in Figure~\ref{fig:vottignasco_predictions} and~\ref{fig:racconigi_predictions}) seem to be lower than the TDC-UnPWaveNet ones. However, this is not true for the Savigliano test predictions, in which the TDC-UnPWaveNet ensemble standard deviation appears to be lower. Furthermore, in our case study, the more variability related to the initialization effect could be also caused by the higher number of parameters and the deeper architecture of the TDC-UnPWaveNet.\\
In terms of performance metrics, our models are in line with, and in some cases even better than, other hydrology DL studies aiming to predict groundwater resources from weather data.
For example, \cite{WunschGroundwater2021} reported mean NSE values around 0.5, and NRMSE between 0.10 and 0.15 for NARX, LSTM, and CNN models.
In \cite{YinComparison2021} authors predicted the groundwater level changes in different locations with data-driven local models, they reported Pearson correlation coefficients (here $\rho$) which are not higher than 0.87. In \cite{LeeUsing2019} a neural network model was adopted to predict groundwater level in South Korea achieving NSE values around 0.8 and $\rho$ about 0.91.\\
What emerges from the present study is that both the TDC-UnPWaveNet and TDC-LSTM have produced satisfactory predictions but with different modeling abilities.
The ability of the TDC-UnPWaveNet to model better the actual temporal dynamics could be due to the more complex structure and the higher number of parameters, which in the framework of DL remain negligible for both models.
Furthermore, it should be considered that the TDC-UnPWaveNet is a CNN-based model, so computationally more efficient thanks to the possibility of parallelization~\cite{BaiEmpirical2018}.

\section{Conclusion}\label{sec:conclusions}

We have proposed two different DL models for predicting, in a many-to-one fashion, the water table depth of three sensors located in the Grana-Maira catchment (Piedmont, IT) from weather image time series. These models are made of two modules: a first Time Distributed CNN (TDC) and a Sequential Module. The TDC is the same for the two proposed models, and it extracts a vectorial representation (Time Distributed Hidden Representation) of the input image time series, i.e. it encodes each image available at each time step into a vector forming a hidden multivariate time series. 
For the TD-LSTM model, the Sequential Module is based on a classical LSTM layer; instead for the TD-UnPWaveNet model the sequential model is based on a new version of the WaveNet adapted here to output a series completely in the future and shorter than the input one - actually a many-to-one scenario in this case study. \\
In developing the UnPWaveNet, and facing the issue of different sequence lengths inside the architecture, we have designed a new Channel Distributed (CD) layer. The CD layer applies the same transformations to each channel individually (i.e. a translation of the concept of Time Distributed layer to channels). In this way, a sequence with many channels could be transformed into a sequence of a different length maintaining the channel-wise dimension. The CD layer, implemented in the UnPWaveNet with a fully connected cell, has proved to be efficient and effective: it achieves very satisfactory results limiting the total number of parameters.\\
Both the DL models have shown remarkable performance, revealing that, in our ROI, it is possible to predict the water table depth using only exogenous weather information with satisfactory results. The TD-LSTM has appeared to be better in terms of bias, but the TD-UnPWaveNet has outperformed the previous in terms of correlation and KGE, appearing to be better in modeling the temporal dynamics of the target. This means that the UnPWaveNet model could be considered as a new possible competitor for recurrent models. Future works are required to investigate better the performance of the UnPWaveNet in other case studies and against other types of DL architecture, e.g. Transformers \cite{VaswaniAttention2017}, here not included because of the already consistent work done in developing and adapting the proposed models to the case study. 

\section{Acknowledgements}

We are grateful for the collaboration with ARPA (Agenzia Regionale Protezione Ambientale) which has been crucial for retrieving all the data for the case study.

\section{Software and data availability}\label{sec:sftw-data}

All the codes and data are available at: \url{https://github.com/Matteo-Salis/water_table_depth_forecasting}.

\section{CRediT}
\textbf{Metteo Salis}: Conceptualization, Methodology, Software \& Data curation, Formal Analysis, Investigation, Writing, Validation, Visualization; \textbf{Abdourrahmane M. Atto}: Conceptualization, Methodology, Supervision, Review, Validation; \textbf{ Stefano Ferraris}: Conceptualization, Supervision, Review, Validation; \textbf{Rosa Meo}: Conceptualization, Methodology, Supervision, Review, Validation. 

\bibliographystyle{elsarticle-num} 
\bibliography{bibliography.bib}




% %% For numbered reference style

%   \textit{\LaTeX: a document preparation system},
%   Addison Wesley, Massachusetts,

% \end{thebibliography}
\end{document}